# Fig Tree-Wasp Symbiotic Coevolutionary Optimization Algorithm

Anand J Kulkarni[1]*; Isha Purnapatre[2]; Apoorva S Shastri[1]

[1]Institute of Artificial Intelligence, Dr Vishwanath Karad MIT World Peace University, 124 Paud Road, Kothrud, Pune 411038, MH, India

[2]Department of Computer Engineering, Cummins College of Engineering for Women, Karvenagar, Pune 411052, MH, India

Email: anand.j.kulkarni@mitwpu.edu.in; ishapurnapatre@gmail.com; apoorva.shastri@mitwpu.edu.in

## Abstract

The nature inspired algorithms are becoming popular due to their simplicity and wider applicability. In the recent past several such algorithms have been developed. They are mainly bio-inspired, swarm based, physics based and socio-inspired; however, the domain based on symbiotic relation between creatures is still to be explored. A novel metaheuristic optimization algorithm referred to as Fig Tree-Wasp Symbiotic Coevolutionary (FWSC) algorithm is proposed. It models the symbiotic coevolutionary relationship between fig trees and wasps. More specifically, the mating of wasps, pollinating the figs, searching for new trees for pollination and wind effect drifting of wasps are modeled in the algorithm. These phenomena help in balancing the two important aspects of exploring the search space efficiently as well as exploit the promising regions. The algorithm is successfully tested on a variety of test problems. The results are compared with existing methods and algorithms. The Wilcoxon Signed Rank Test and Friedman Test are applied for the statistical validation of the algorithm performance. The algorithm is also further applied to solve the real-world engineering problems. The performance of the FWSC underscored that the algorithm can be applied to wider variety of real-world problems.

## 1. Introduction

Several deterministic and approximation-based optimization algorithms have been developed so far. The approximation-based algorithms are majorly inspired from nature. These algorithms are generally classified as bio-inspired, swarm based, socio-inspired and physics based (Kumar et al, 2018). The bio-inspired algorithms are generally inspired from the biological evolution of the individuals such as Genetic Algorithms (GAs) (Holland J.H., Katoch 2021, Differential Evolution (DE) (Price 2013), Snail Homing and Mating Search Algorithm (SHMS) (Kulkarni et al, 2024), Artificial Immune System (Timmis 2004), etc. Particle Swarm Optimization (PSO) Kennedy & Eberhart 1995), Artificial Bee Colony (ABC) (Karaboga & Basturk 2007), Salp Swarm Algorithm SSA (Mirjalili et al 2017), Ant Colony Optimization (ACO) (Dorigo & Stutzle 2019), Krill Herd (KH) (Gandomi & Alavi 2012) are few of the Swarm based methods. The Socio-inspired optimization algorithms are Cohort Intelligence, Political Optimizer (Askari et al 2020), Ideology Algorithm (Huan et al 2017), Socio Evolution & Learning Optimization (SELO) Algorithm (Kumar et al, 2018), League Championship Algorithm (LCA) (Kashan 2014), Teaching and Learning based Optimization algorithm (TLBO) (Rao et al 2011) are to mention a few.

There are certain algorithms which are based on symbiotic relations of the living beings. A population-based approach inspired from the mutualism, commensalism and parasitism referred to as Symbiotic Organisms Search (SOS) was developed by Cheng and Prayago (2024). According to The et al 2020, the SOS algorithm lacks the balance between exploration and exploitation due to absence of hyper parameter tuning of the algorithm. Miao et al (2020) developed Phasor SOS (PSOS) helping the algorithm jump out of local solutions with improved robustness. In addition, Prayago et al (2017), proposed Enhanced SOS (ESOS) with added cleptoparasitism approach for expanded search space which eventually helped in improving exploration ability (Askarzadeh 2016). Furthermore, a Multi-Agent-based Symbiotic Organism Search (MASOS) as proposed by Kawambwa & Mnyanghwalo (2023) for solving the problems in distributed way avoiding single point failure. The Binary SOS was developed by Du et al (2020). Li et al (2023) developed hybridized Whale Optimization Algorithm by incorporating improved SOS as a search strategy. The exploration and exploitation strategies of the SOS were improved by incorporating levy flight and spiral updating approach, respectively. The symbiotic coevolution of several living species exists. The notable examples are crocodile-and-plover wading behavior, Coyotes-and-badgers hunting behavior, ants-and-aphids mutualistic relationship, etc are to mention a few. The work here proposes a novel metaheuristic referred to as Fig Tree-Wasp Symbiotic Coevolutionary (FWSC) algorithm. It models symbiotic and coevolutionary behavior of fig trees and wasps. It is important to mention here that the fig tree allow the pollination from the wasps of the superfamily Chalcidoidea (Dunn 2020). The opening of the figs is also evolved such that it allows only these wasps to enter. The wasps are also evolved to search for these figs, enter inside the figs and pollinate there. The FWSC algorithm is tested on 23 benchmark test functions and also applied for solving real world constrained problems such as Pressure Vessel Design, Stepped Beam Design and Welded Beam Design. The results have been compared with the existing contemporary algorithms.

The manuscript is organized as follows: Section 2 provides biological description of the FWSC algorithm including mathematical formulation. The validation of the algorithm by solving the test problems is elaborated in Section 3. Solution to the real-world problems is provided in Section 4. The conclusions and future directions are discussed in Section 5 of the manuscript.

## 2. FWSC

A fig is essentially a garden of flowers. The wasps are the only pollinators of these fig trees which allow them to coevolve. The tiny (not more than $2mm$ in length) female wasps loaded with pollens enter the fig through its opening. The entry hole is too small which makes them loose their wings. These female wasps crawl inside the fig and lay their eggs in the flowers as well as pollinate them. They die inside the fig and are further digested by the fig itself. Within approximately two weeks' time, the fig ripens and first emerge the male wasps from their eggs. They mate with the females which are still in their eggs. The males then make a hole in fig and become prey to the outside insects and birds giving ultimate sacrifice. Then the pregnant females loaded with the pollens collected from the fig fly out and search for another fig tree for pollination. This makes both the fig trees and wasps symbiotically coevolve. While searching the fig tree, few of them are drifted randomly to a longer distance. This randomness helps the FWSC algorithm jumping out of local minima. Furthermore, mating between the male and female wasps generates offsprings from within the promising region combining the characteristics of the parents. The further search direction of the offsprings is decided based on the extreme limits generated while locating the promising region. It necessary drives the search space exploration by the wasps and generate the locations where a fig tree can be pollinated. This helps in finding the promising figs with better solutions reaching out to fig trees with promising locations, more specifically the fig tree solutions meet with the fig tree solutions generating better ones together. The natural cycle of the FWSC is presented in Figure 1 and the algorithmic process is explained in this section along with the flowchart in Figure 2.

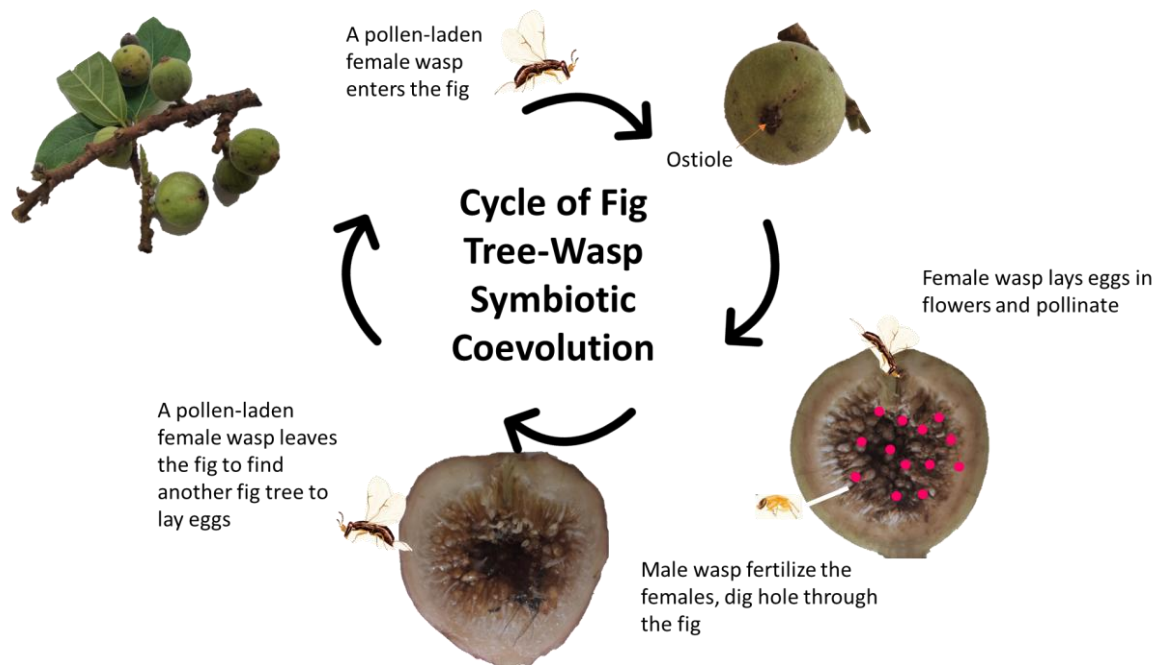

Figure 1: Cycle of Fig Tree-Wasp Symbiotic Coevolution

The graphical representation of Figs and Wasp is shown in Figure 1. Here, $t$ ($t = 1,2 … … T$) denote the number of trees while $a$ ($a = 1,2 … … A$) denote the number of figs and similarly $w$ ($w = 1,2, … … W$) denote the number of wasps. Each fig has some parent wasps (denoted by dots) and their offspring wasps (denoted by cross).

Initialize the iteration number $k = 1$, neighborhood constant $\eta_0$, wind effect threshold $\theta$, number of fig trees $T$, number of figs per tree $a^t$ wasps per tree $t = 1, …, T$.

Assume the function $f$ being minimized represents the forest, i.e.,

$$minimize$$

$$f(\boldsymbol{x}) = f(x_1, …, x_i, …, x_n) \tag{1}$$

$$\forall_i \ x_i^l \leq x_i \leq x_i^u$$

**Step 1: Generate fig trees, figs and wasps**

Generate $T$ trees in the forest as follows:

$$\forall_{t,i} t^k = [x_l^i + rand(x_l^u - x_l^i)] \pm \eta_t^k \qquad (2)$$

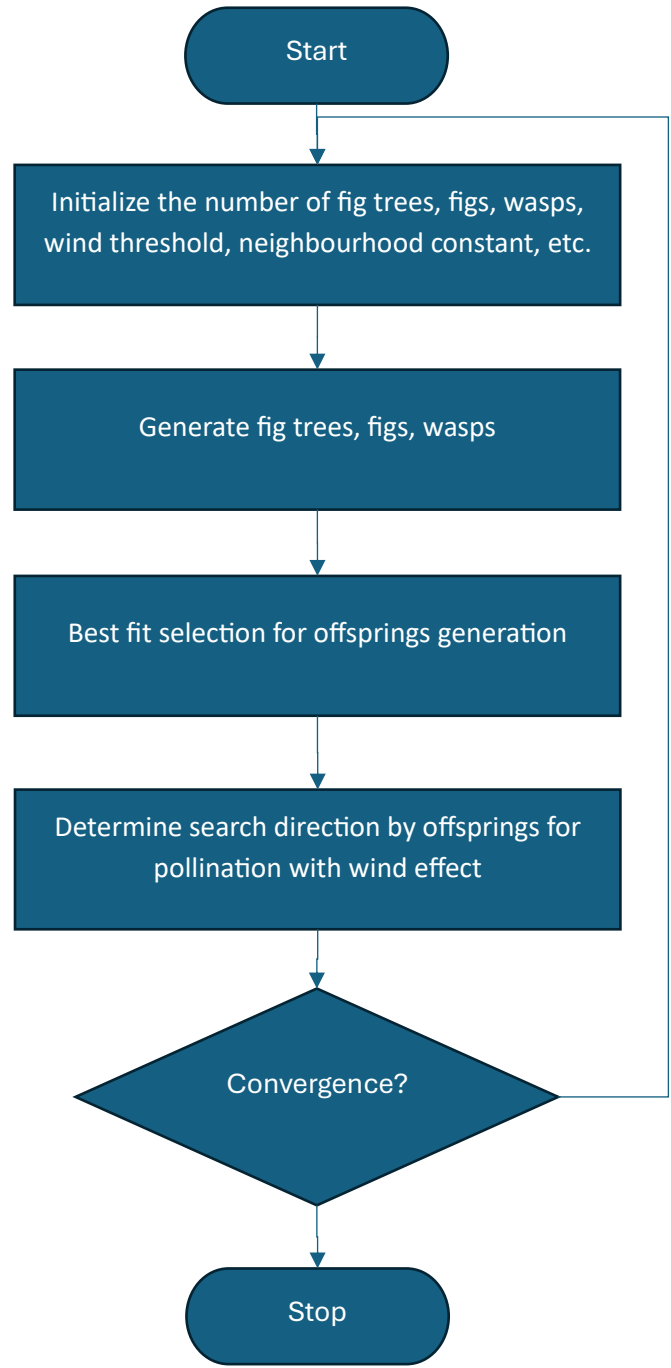

**Figure 2: Flowchart of Fig Tree-Wasp Symbiotic Coevolutionary Algorithm**

where $\eta_t^k = \eta_0 e(1 - \frac{k}{K})$ is the neighborhood region of every fig tree $t$ and $K$ is a very high number.

Generate figs $a^t$ in every fig tree $t$ as follows:

$$\forall_t a^t = [t^l + rand(t^u - t^l)] \pm \eta_t^k \tag{3}$$

Generate wasps $w^{a,t}$ For every fig tree $t$ and associated fig $a$ as follows:

$$w^{a,t} = [a^{t,l} + rand(a^u - a^l)] \tag{4}$$

**Step 2: Best fit Selection for offsprings generation**

Evaluate every wasp fitness based on its function value $\forall t, a \ f_{w_a^t}(x)$. For every tree $t$ and associated fig $a$, randomly segregate $\frac{W_a^t}{2}$ wasps as females and the rest $\frac{W_a^t}{2}$ as males.

Apply Best Fit Selection method for wasps mating as follows:

**2.1:** Arrange $\frac{W_a^t}{2}$ female wasps in ascending order based on their respective function value, i.e. $f_{w_a^t}^{fl1}(x) \leq f_{w_a^t}^{fl2}(x) \leq \ldots f_{w_a^t}^{flr}(x) \ldots \leq f_{w_a^t}^{flR}(x)$ , $R = \frac{W_a^t}{2}$ and generate a grid of cumulative function values as follows:

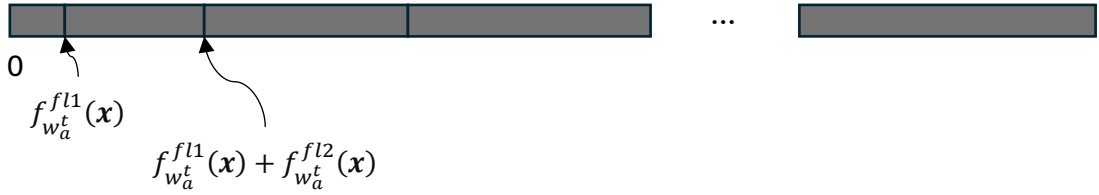

$f_{w_a^t}^{fl1}(x)$

$f_{w_a^t}^{fl1}(x) + f_{w_a^t}^{fl2}(x)$

$$f_{w_a^t}^{fl1}(x) \leq f_{w_a^t}^{fl2}(x) \leq \ldots f_{w_a^t}^{flr}(x) \ldots \leq f_{w_a^t}^{flR}(x) \ , R = \frac{W_a^t}{2}$$

**3:** All the $\frac{W_a^t}{2}$ male wasps locate their function values from within the grid defined. If $\forall j \ f_{w_a^t}^{flr}(x) \leq f_{w_a^t}^{mlj}(x) \leq f_{w_a^t}^{flr+1}(x)$ , $j = 1, \ldots, \frac{W_a^t}{2}$, then an offspring variables $x_{w_a^t}^o$ are generated based on the average values of the variables of the mating wasp as follows:

$$\forall a, t, j \ \ x_{w_a^t}^o = \left(\frac{x_{w_a^t}^{flr} + x_{w_a^t}^{flr+1}}{2}\right) \ , o = 1, \ldots, \frac{W_a^t}{2} \tag{5}$$

The pool of offspring variables is represented as follows:

$$P = \begin{bmatrix} x^1 & \cdots & x^o & \cdots & x^{\frac{W_a^t}{2}} \end{bmatrix} \tag{6}$$

**Step 3: Determine Search direction by Offsprings for Pollination**

Every offspring $o$ , $o = 1, \ldots, \frac{W_a^t}{2}$ decides q search direction based on the minimum and maximum variable values located from within the pool $P$.

$$\forall o, i \ \ \ x_i^o = min \ x_i^p + rand(min \ x_i^p - min \ x_i^p) \tag{7}$$

**Step 4: Wind Effect**

If $(rand(0,1) \leq \theta)$ then 10% of the wasps are perturbed as follows:

$$x_i^o = x_i^o + x_i^o \times rand(0,1) \qquad (8)$$

**Step 5: Generate new trees, figs and wasps**

The best $T$ offsprings are selected from within the pool $P$. Update $k = k + 1$, and $\eta_t^k = \eta_0 e(1 - \frac{k}{K})$. As discussed in Step 1, generate Figs $a^t$ in every fig tree as follows: $t\ \forall_t a^t = [t^l + rand(t^u - t^l)] \pm \eta_t^k$ and for every fig tree $t$ and associated fig $a$ generate wasps $w^{a,t}$ as follows: $w^{a,t} = [a^{t,l} + rand(a^u - a^l)]$

If converged, then stop, else continue to Step 2.

## 3. Solution to Benchmark Test Functions

The algorithm's effectiveness is tested by applying it to a variety of benchmark test functions, as outlined by Abdollahzadeh et al. (2021). These benchmark functions are categorized into two groups: unimodal (UM) and multi-modal (MM) functions. Detailed information about the function names and characteristics for both UM and MM categories is presented in Tables 1-3. To evaluate the performance, each problem in these benchmark sets is solved 30 times using the FWSC algorithm. The algorithm is coded in Python 3.10.7 with Intel(R), Core(TM) i7-8565U CPU, 1.80GHz and 16GB RAM. The algorithmic parameters such as number of fig trees $T = 3$, number of figs per tree $a^t = 4$, neighborhood constant $\eta_0 = 0.8$ and number of wasps per fig $w^{a,t} = 8$.

**Table 1 Details of UM benchmark functions**

| No. | Type | Function | Dimensions | Range | $F_{min}$ |
|---|---|---|---|---|---|
| F1 | US | Sphere | 30, 100, 500, 1000 | $[-100, 100]^d$ | 0 |
| F2 | UN | Schwefel 2.22 | 30, 100, 500, 1000 | $[-10, 10]^d$ | 0 |
| F3 | UN | Schwefel 1.2 | 30, 100, 500, 1000 | $[-100, 100]^d$ | 0 |
| F4 | US | Schwefel 2.21 | 30, 100, 500, 1000 | $[-100, 100]^d$ | 0 |
| F5 | UN | Rosenbrock | 30, 100, 500, 1000 | $[-30, 30]^d$ | 0 |
| F6 | US | Step | 30, 100, 500, 1000 | $[-100, 100]^d$ | 0 |
| F7 | US | Quartic | 30, 100, 500, 1000 | $[-128, 128]^d$ | 0 |

**Table 2 Details of MM benchmark functions**

| No. | Type | Function | Dimensions | Range | $F_{min}$ |
|---|---|---|---|---|---|
| F8 | MS | Schwefel | 30, 100, 500, 1000 | $[-500, 500]^d$ | $-418.9829 \times n$ |
| F9 | MS | Rastrigin | 30, 100, 500, 1000 | $[-5.12, 5.12]^d$ | 0 |
| F10 | MN | Ackley | 30, 100, 500, 1000 | $[-32, 32]^d$ | 0 |
| F11 | MN | Griewank | 30, 100, 500, 1000 | $[-600, 600]^d$ | 0 |
| F12 | MN | Penalized | 30, 100, 500, 1000 | $[-50, 50]^d$ | 0 |
| F13 | MN | Penalized2 | 30, 100, 500, 1000 | $[-50, 50]^d$ | 0 |

Table 3 Details of fixed-dimension MM benchmark functions

| No. | Type | Function | Dimensions | Range | $F_{min}$ |
|---|---|---|---|---|---|
| F14 | FM | Foxholes | 2 | $[-65, 65]^d$ | 1 |
| F15 | FM | Kowalik | 4 | $[-5, 5]^d$ | 0.0003 |
| F16 | FM | Six Hump Camel | 2 | $[-5, 5]^d$ | -1.0316 |
| F17 | FM | Branin | 2 | $[-5, 5]^d$ | 0.398 |
| F18 | FM | Goldstein-Price | 2 | $[-2, 2]^d$ | 3 |
| F19 | FM | Hartman 3 | 3 | $[1, 3]^d$ | -3.86 |
| F20 | FM | Hartman 6 | 6 | $[0, 1]^d$ | -3.32 |
| F21 | FM | Shekel 5 | 4 | $[0, 10]^d$ | -10.1532 |
| F22 | FM | Shekel 7 | 4 | $[0, 10]^d$ | -10.4028 |
| F23 | FM | Shekel 10 | 4 | $[0, 10]^d$ | -10.5363 |

The symbiotic coevolution between fig trees and wasps modeled in the FWSC algorithm is represented in Figure 3 using Ackley function as a representative problem. Figures 3 (a–c) depict the pollinating wasps, while Figures 3 (d–f) illustrate the corresponding evolution of trees. Initially, the trees are randomly distributed within the search space, representing a diverse initial population. As the wasps emerge, they traverse the search space, identifying and selecting the better trees for pollination. This selection process ensures that each successive generation of trees is positioned in increasingly favorable regions of the search space, thereby enhancing their fitness. Concurrently, the wasps also undergo an evolution through mating at different tree locations improving their ability to locate and pollinate trees improves over time. As the trees evolve towards optimal positions, the wasps that emerge from them inherit advantageous traits of finding more optimal trees which enhances their search efficiency. This reciprocal enhancement between the trees and wasps' results in an iterative refinement process, where both species coevolve. The convergence of both trees and wasps towards optimal solutions in the search space serves as empirical evidence of their coevolutionary behavior, which is represented in Figure 3.

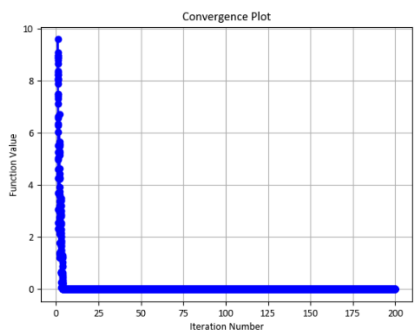
(a)

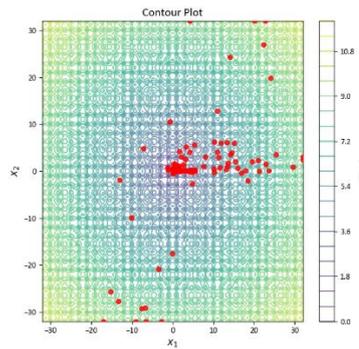
(b)

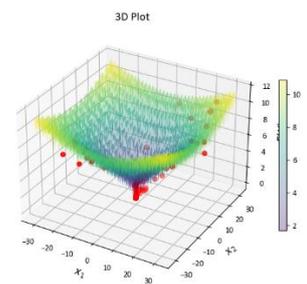
(c)

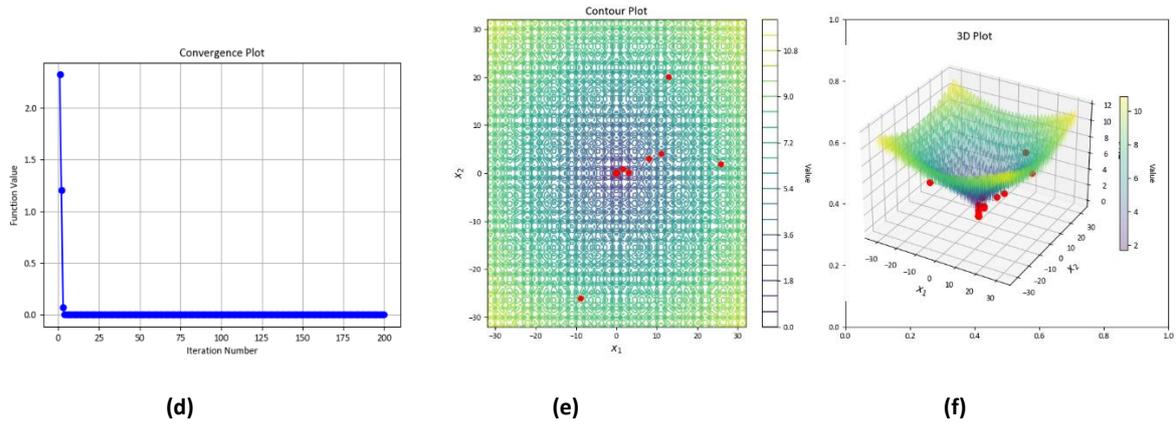

Figure 3: Symbiotic Coevolution of Trees and Wasps

The proposed FWSC algorithm is evaluated by comparing its performance with various well-known algorithms: African Vultures Optimization Algorithm (AVOA) (Abdollahzadeh et al., 2021), PSO (Simon, 2008), Grey Wolf Optimizer (GWO) (Mirjalili et al., 2014), Farmland Fertility Algorithm (FFA) (Shayanfar & Gharehchopogh, 2018), Whale Optimization Algorithm (WOA) (Mirjalili & Lewis, 2016), Teaching–Learning-Based Optimization (TLBO) (Rao, Savsani, & Vakharia, 2012), Moth-Flame Optimization (MFO) (Mirjalili, 2015), Biogeography-Based Optimization (BBO) (Simon, 2008), Differential Evolution (DE) (Simon, 2008), Salp Swarm Algorithm (SSA) (Mirjalili et al., 2017), Gravitational Search Algorithm (GSA) (Rashedi et al., 2009), Inclined Planes System Optimization (IPO) (Mozaffari et al., 2016), and the Snail Homing and Mating Search Algorithm (SHMS) (Kulkarni, 2023).

The evaluation examines the mean, best solution, and standard deviation (STD) over 30 independent runs of the FWSC algorithm when solving the F1–F13 benchmark problems across four different dimensions: 30, 100, 500, and 1000. The corresponding results are presented in Tables 4–7. For a more in-depth analysis, Table 8 presents a comparison of FWSC's performance on fixed-dimension MM benchmark problems (F14–F23). To better understand the wasps' optimization behaviour, graphical representations such as convergence plots, contour plots, and 3D mesh plots are provided. Figures 4(a)–4(g) illustrate FWSC's performance on UM benchmark functions, while Figures 4(h)–4(m) present its results on MM benchmark functions. Additionally, Figures 5(a)–5(d) showcase the algorithm's performance on fixed-dimension MM benchmark functions. From Table 4-7, it is evident that FWSC algorithm outperforms other algorithms significantly except AVOA. In addition, it shows the ability to solve problems having larger dimensions with same set of parameters and without affecting its performance. Table 8 shows the performance comparison of FWSC algorithm for fixed dimension MM problems F14-F23. Except few, the FWSC algorithm yields promising results for these MM functions. However, FWSC algorithm is not efficient in solving some of the test functions as it is evident that it trapped in local minima.

The statistical comparison between FWSC and the competing algorithms is conducted using the Wilcoxon signed-rank test. The results of this pairwise statistical comparison, showing which algorithms performed better in solving F1–F13 for different dimensions (30, 100, 500, and 1000), are summarized in Tables 14–17. The Wilcoxon signed-rank test highlighted that FWSC algorithm outperformed most of the other algorithms, even with increasing dimensions. Table 18 shows pairwise comparison of F14-F23 problems where FWSC algorithm could not yield better solution compared to other algorithms. Additionally, the Friedman test was performed to ensure statistical validity. This nonparametric test accounts for multiple comparisons and ranks the algorithms based on their performance. It calculates the average of the optimal solutions obtained from 30 runs of FWSC solving F1–F13 and F14–F23 problems and compares them against other optimization algorithms on the same benchmark test functions. The ranking results from the Friedman test, based on best values, are presented in Tables 9–13.

**Table 4 Results of benchmark functions (F1-F13), with 30 dimensions**

| No. | | AVOA | PSO | GWO | FFA | WOA | TLBO | MFO | BBO | DE | SSA | GSA | IPO | SHMS | **FWSC** |
|---|---|---|---|---|---|---|---|---|---|---|---|---|---|---|---|
| F1 | Best | 5.44E-269 | 1.37E+03 | 6.81E-29 | 2.27E-07 | 2.43E-87 | 1.97E-100 | 1.01E-01 | 3.15E+00 | 7.46E-05 | 2.75E-08 | 2.11E-16 | 1.82E-12 | 0.00E+00 | **0.00E+00** |
| | Worst | 6.05E-198 | 4.27E+03 | 3.14E-26 | 2.17E-06 | 2.01E-69 | 2.26E-95 | 1.00E+04 | 6.33E+00 | 7.89E-04 | 8.75E-07 | 8.65E-03 | 3.88E-10 | 0.00E+00 | **0.00E+00** |
| | Mean | 2.01,199 | 2.37E+03 | 2.59E-27 | 7.88E-07 | 8.04E-71 | 1.66E-96 | 1.66E+03 | 4.47E+00 | 3.36E-04 | 2.07E-07 | 2.88E-04 | 4.76E-11 | 0.00E+00 | **0.00E+00** |
| | STD | 0.00E+00 | 6.17E+02 | 5.83E-27 | 5.15E-07 | 3.72E-70 | 4.36E-96 | 3.78E+03 | 7.54E-01 | 1.83E-04 | 2.17,07 | 1.58E-03 | 8.42E-11 | 0.00E+00 | **0.00E+00** |
| F2 | Best | 2.87E-136 | 1.13E+01 | 1.29E-17 | 1.35E-06 | 5.07E-59 | 1.57E-50 | 8.00E-02 | 3.90E-01 | 1.54E-03 | 2.55E-01 | 5.64E-08 | 6.77E-08 | 0.00E+00 | **3.75E-41** |
| | Worst | 2.58E-102 | 2.77E+01 | 4.34E-16 | 8.00E-06 | 2.16E-48 | 1.11E-48 | 7.00E+01 | 6.00E-01 | 5.95E-03 | 8.00E+00 | 1.19E+00 | 1.27E-05 | 0.00E+00 | **1.11E-36** |
| | Mean | 8.72E-104 | 1.91E+01 | 9.02E-17 | 3.82E-06 | 7.41E-50 | 1.89E-49 | 2.92E+01 | 5.05E-01 | 3.48E-03 | 2.39E+00 | 1.40E-01 | 8.28E-07 | 0.00E+00 | **4.08E-38** |
| | SID | 4.71E-103 | 3.55E+00 | 8.98E-17 | 1.54E-06 | 3.94E-49 | 2.79E-49 | 2.03E+01 | 4.53E-02 | 1.00E-03 | 1.72E+00 | 3.34E-01 | 2.37E-06 | 0.00E+00 | **2.03E-37** |
| F3 | Best | 4.92E-217 | 4.07E+03 | 9.78E-08 | 2.78E+03 | 1.02E+04 | 2.76E-26 | 3.45E+03 | 2.51E+02 | 2.17E+04 | 4.40E+02 | 6.57E+02 | 1.30E+00 | 0.00E+00 | **3.54E-35** |
| | Worst | 2.35E-143 | 2.29E+04 | 7.58E-04 | 7.39E+03 | 6.08E+04 | 2.93E-21 | 4.48E+04 | 1.00E+03 | 5.36E+04 | 5.37E+03 | 1.77E+03 | 4.53E+00 | 0.00E+00 | **1.57E-30** |
| | Mean | 7.83E-145 | 9.20E+03 | 4.02E-05 | 5.03E+03 | 4.18E+04 | 3.35E-22 | 2.23E+04 | 4.98E+02 | 3.64E+04 | 1.83E+03 | 1.09E+03 | 2.41E+00 | 0.00E+00 | **2.04E-31** |
| | STD | 4.29E-144 | 4.24E+03 | 1.46E-04 | 1.32E+03 | 1.32E+04 | 6.76E-22 | 1.19E+04 | 1.49E+02 | 7.01E+03 | 1.02E+03 | 3.11E+02 | 7.86E-01 | 0.00E+00 | **3.38E-31** |
| F4 | Best | 2.57E-134 | 1.57E+01 | 6.93E-08 | 1.33E+01 | 5.27E+00 | 1.02E-42 | 5.39E+01 | 1.21E+00 | 6.27E+00 | 5.22E+00 | 4.27E+00 | 2.41E-02 | 0.00E+00 | **4.27E-34** |
| | Worst | 2.84E-102 | 2.80E+01 | 6.17E-06 | 2.11E+01 | 8.61E+01 | 3.41E-40 | 8.28E+01 | 2.12E+00 | 1.75E+01 | 1.52E+01 | 1.12E+01 | 8.47E-02 | 0.00E+00 | **1.56E-30** |
| | Mean | 1.03E-103 | 2.14E+01 | 7.54E-07 | 1.64E+01 | 4.94E+01 | 9.51E-41 | 6.77E+01 | 1.56E+00 | 9.40E+00 | 1.06E+01 | 7.43E+00 | 4.59E-02 | 0.00E+00 | **2.62E-31** |
| | STD | 5.18E-103 | 3.51E+00 | 1.16E-06 | 2.39E+00 | 2.79E+01 | 1.07E-40 | 8.29E+00 | 2.00E-01 | 2.32E+00 | 2.54E+00 | 1.95E+00 | 1.48E-02 | 0.00E+00 | **3.86E-31** |
| F5 | Best | 1.96E-05 | 1.48E+05 | 2.61E+01 | 1.51E+01 | 2.72E+01 | 2.66E+01 | 1.17E+02 | 4.99E+01 | 2.75E+01 | 2.51E+01 | 2.46E+01 | 1.34E+02 | 2.89E+01 | **2.81E+01** |
| | Worst | 2.56E-02 | 1.62E+06 | 2.88E+01 | 1.85E+02 | 2.84E+01 | 2.84E+01 | 7.99E+07 | 1.64E+03 | 2.38E+02 | 1.08E+03 | 3.27E+02 | 3.96E+02 | 2.90E+01 | **2.90E+01** |
| | Mean | 6.50E-03 | 5.15E+05 | 2.70E+01 | 7.31E+01 | 2.76E+01 | 2.72E+01 | 2.69E+06 | 2.24E+02 | 6.00E+01 | 1.85E+02 | 8.34E+01 | 2.27E+02 | 2.90E+01 | **2.87E+01** |
| | STD | 7.15E-03 | 3.37E+05 | 7.53E-01 | 3.94E+01 | 3.94E-01 | 4.23E-01 | 1.45E+07 | 3.60E+02 | 6.31E+01 | 2.46E+02 | 6.72E+01 | 7.44E+01 | 2.24E-02 | **3.18E-01** |
| F6 | Best | 2.43E-07 | 1.19E+03 | 9.91E-05 | 1.07E-06 | 1.00,01 | 3.37E-07 | 2.80,01 | 1.37E+00 | 7.32E-05 | 8.02,08 | 1.16E-16 | 0.00E+00 | 0.00E+00 | **0.00E+00** |
| | Worst | 6.58E-06 | 3.55E+03 | 1.47E+00 | 1.24E-01 | 8.78E-01 | 5.71E-05 | 1.95E+04 | 3.17E+00 | 5.72E-04 | 6.62E-05 | 5.48E-03 | 4.00E+00 | 0.00E+00 | **0.00E+00** |
| | Mean | 2.43E-06 | 2.35E+03 | 7.34E-01 | 4.26E-03 | 3.80E-01 | 1.06E-05 | 2.66E+03 | 2.36E+00 | 2.65E-04 | 9.76E-06 | 1.83E-04 | 9.33E-01 | 0.00E+00 | **0.00E+00** |
| | STD | 1.89E-06 | 5.34E+02 | 3.06E-01 | 2.25E-02 | 1.91E-01 | 1.28E-05 | 5.18E+03 | 4.18,01 | 1.47E-04 | 1.61E-05 | 1.00E-03 | 1.17E+00 | 0.00E+00 | **0.00E+00** |
| F7 | Best | 4.26E-06 | 1.00,01 | 6.31E-04 | 1.94E-02 | 1.60E-04 | 4.27E-04 | 9.10,02 | 6.34E-03 | 3.00E-02 | 6.10,02 | 2.22E-02 | 9.85E-03 | 3.27E-04 | **1.38E-142** |
| | Worst | 7.61E-04 | 8.91E-01 | 6.15E-03 | 8.19E-02 | 1.46E-02 | 2.66E-03 | 8.25E+00 | 2.67E-02 | 7.44E-02 | 3.17E-01 | 1.60E-01 | 4.36E-02 | 1.62E-02 | **3.17E-122** |
| | Mean | 2.52E-04 | 3.77E-01 | 2.35E-03 | 5.03E-02 | 3.24E-03 | 1.27E-03 | 1.33E+00 | 1.62E-02 | 4.90E-02 | 1.71E-01 | 8.69E-02 | 2.75E-02 | 4.88E-03 | **2.47E-123** |
| | STD | 2.05E-04 | 2.10E-01 | 1.34E-03 | 1.51E-02 | 3.45E-03 | 5.77E-04 | 2.58E+00 | 5.71E-03 | 1.15E-02 | 7.63E-02 | 3.90E-02 | 9.80E-03 | 3.53E-03 | **7.61E-123** |

| No. | | AVOA | PSO | GWO | FFA | WOA | TLBO | MFO | BBO | DE | SSA | GSA | IPO | SHMS | **FWSC** |
|---|---|---|---|---|---|---|---|---|---|---|---|---|---|---|---|
| F8 | Best | -1.25E+04 | -5.10E+03 | -7.31E+03 | -9.37E+03 | -1.25E+04 | -9.54E+03 | -1.00E+04 | -9.04E+03 | -7.59E+03 | -9.33E+03 | -3.34E+03 | -4.12E+03 | 0.00E+00 | **1.18E+04** |
| | Worst | -1.21E+04 | -2.69E+03 | -4.69E+03 | -4.88E+03 | -7.98E+03 | -5.86E+03 | -6.47E+03 | -7.01E+03 | -6.05E+03 | -6.14E+03 | -1.70E+03 | -2.52E+03 | 0.00E+00 | **1.23E+04** |
| | Mean | -1.25E+04 | -3.82E+03 | -5.99E+03 | -6.89E+03 | -9.69E+03 | -7.39E+03 | -8.51E+03 | -8.10E+03 | -6.72E+03 | -7.32E+03 | -2.47E+03 | -3.28E+03 | 0.00E+00 | **1.21E+04** |
| | STD | 1.00E+02 | 6.19E+02 | 6.14E+02 | 1.14E+03 | 1.50E+03 | 9.60E+02 | 7.76E+02 | 5.65E+02 | 3.51E+02 | 7.84E+02 | 3.72E+02 | 3.60E+02 | 0.00E+00 | **1.25E+02** |
| F9 | Best | 0.00E+00 | 1.08E+02 | 5.68E-14 | 3.59E+01 | 0.00E+00 | 0.00E+00 | 7.57E+01 | 2.59E+01 | 1.29E+02 | 2.98E+01 | 1.39E+01 | 9.25E+00 | 0.00E+00 | **0.00E+00** |
| | Worst | 0.00E+00 | 2.11E+02 | 1.93E+01 | 1.66E+02 | 0.00E+00 | 4.52E+01 | 2.61E+02 | 8.01E+01 | 1.74E+02 | 9.55E+01 | 4.68E+01 | 3.17E+01 | 0.00E+00 | **0.00E+00** |
| | Mean | 0.00E+00 | 1.48E+02 | 2.00E+00 | 7.96E+01 | 0.00E+00 | 2.08E+01 | 1.55E+02 | 5.22E+01 | 1.57E+02 | 5.05E+01 | 2.92E+01 | 1.68E+01 | 0.00E+00 | **0.00E+00** |
| | STD | 0.00E+00 | 2.16E+01 | 4.07E+00 | 3.10E+01 | 0.00E+00 | 9.68E+00 | 4.15E+01 | 1.30E+01 | 1.06E+01 | 1.58E+01 | 7.25E+00 | 4.32E+00 | 0.00E+00 | **0.00E+00** |
| F10 | Best | 8.88E-16 | 6.71E+00 | 7.19E-14 | 9.29E-05 | 8.84E-16 | 4.44E-15 | 5.58E-01 | 3.80E-01 | 3.04E-03 | 1.34E+00 | 7.68E-09 | 1.61E+00 | 4.44E-16 | **4.44E-16** |
| | Worst | 8.88E-16 | 1.20E+01 | 1.35E-13 | 8.73E-04 | 7.95E-15 | 7.99E-15 | 2.00E+01 | 7.63E-01 | 7.78E-03 | 6.28E+00 | 1.16E+00 | 3.83E+00 | 4.44E-16 | **4.44E-16** |
| | Mean | 8.88E-16 | 1.02E+01 | 9.70E-14 | 3.44E-04 | 4.08E-15 | 5.98E-15 | 1.40E+01 | 5.90E-01 | 5.19E-03 | 2.99E+00 | 6.95E-02 | 2.30E+00 | 4.44E-16 | **4.44E-16** |
| | STD | 0.00E+00 | 1.30E+00 | 1.55E-14 | 1.62E-04 | 2.35E-15 | 1.79E-15 | 7.67E+00 | 9.94E-02 | 1.32E-03 | 1.19E+00 | 2.66E-01 | 4.82E-01 | 0.00E+00 | **0.00E+00** |
| F11 | Best | 0.00E+00 | 1.22E+01 | 0.00E+00 | 2.77E-06 | 0.00E+00 | 0.00E+00 | 6.32E-01 | 9.32E-01 | 3.26E-04 | 1.59E-03 | 1.21E+01 | 1.02E-03 | 0.00E+00 | **0.00E+00** |
| | Worst | 0.00E+00 | 3.24E+01 | 2.93E-02 | 3.53E-02 | 1.89E-01 | 1.20E-04 | 9.10E+01 | 1.07E+00 | 1.29E-01 | 6.12E-02 | 4.75E+01 | 4.30E-02 | 0.00E+00 | **0.00E+00** |
| | Mean | 0.00E+00 | 2.11E+01 | 4.70E-03 | 6.41E-03 | 2.21E-02 | 4.98E-06 | 8.37E+00 | 1.01E+00 | 1.44E-02 | 1.69E-02 | 2.88E+01 | 1.10E-02 | 0.00E+00 | **0.00E+00** |
| | STD | 0.00E+00 | 5.28E+00 | 8.48E-03 | 7.85E-03 | 5.38E-02 | 2.23E-05 | 2.34E+01 | 2.46E-02 | 3.48E-02 | 1.26E-02 | 6.76E+00 | 9.04E-03 | 0.00E+00 | **0.00E+00** |
| F12 | Best | 1.72E-08 | 9.28E+00 | 1.26E-02 | 7.85E-05 | 5.89E-03 | 6.03E-10 | 2.08E+00 | 2.94E-03 | 5.02E-05 | 1.44E+00 | 1.04E-01 | 3.05E-02 | 0.00E+00 | **6.76E-01** |
| | Worst | 1.29E-06 | 3.58E+03 | 1.15E-01 | 1.96E+00 | 3.17E-01 | 1.81E-05 | 5.04E+03 | 1.38E-02 | 1.46E-03 | 1.43E+01 | 4.34E+00 | 1.46E+00 | 0.00E+00 | **1.41E+00** |
| | Mean | 3.79E-07 | 2.74E+02 | 5.22E-02 | 1.32E-01 | 2.95E-02 | 1.25E-06 | 1.80E+02 | 8.11E-03 | 3.92E-04 | 7.24E+00 | 2.01E+00 | 3.75E-01 | 0.00E+00 | **1.06E+00** |
| | STD | 3.32E-07 | 6.87E+02 | 2.56E-02 | 3.72E-01 | 5.49E-02 | 3.53E-06 | 9.18E+02 | 2.47E-03 | 4.13E-04 | 3.26E+00 | 1.13E+00 | 3.69E-01 | 0.00E+00 | **1.52E-01** |
| F13 | Best | 8.67E-07 | 1.36E+04 | 6.35E-02 | 1.01E-04 | 1.44E-01 | 4.00E-02 | 4.14E+00 | 8.53E-02 | 2.20E-04 | 2.09E-02 | 4.00E-02 | 6.38E-02 | 1.74E+00 | **4.15E-05** |
| | Worst | 8.85E-05 | 9.43E+05 | 1.05E+00 | 2.89E-01 | 1.37E+00 | 1.01E+00 | 1.91E+03 | 1.80E-01 | 7.14E-03 | 4.88E+01 | 3.69E+01 | 2.71E-01 | 3.00E+00 | **2.19E-01** |
| | Mean | 1.10E-05 | 1.66E+05 | 6.10E-01 | 4.78E-02 | 5.38E-01 | 4.38E-01 | 1.24E+02 | 1.23E-01 | 1.60E-03 | 1.80E+01 | 1.17E+01 | 1.21E-01 | 2.76E+00 | **1.04E-01** |
| | STD | 1.56E-05 | 2.15E+05 | 2.63E-01 | 7.61E-02 | 3.19E-01 | 2.69E-01 | 3.82E+02 | 2.52E-02 | 1.59E-03 | 1.41E+01 | 8.46E+00 | 4.09E-02 | 3.11E-01 | **5.94E-02** |

**Table 5 Results of benchmark functions (F1-F13), with 100 dimensions**

| No. | | AVOA | PSO | GWO | FFA | WOA | TLBO | MFO | BBO | DE | SSA | GSA | IPO | SHMS | **FWSC** |
|---|---|---|---|---|---|---|---|---|---|---|---|---|---|---|---|
| F1 | Best | 4.37E-260 | 9.85E+03 | 3.22E-13 | 1.75E+03 | 7.24E-84 | 2.00E-92 | 3.00E+04 | 1.66E+02 | 2.44E+03 | 7.65E+02 | 2.74E+03 | 3.59E-01 | 0.00E+00 | **0.00E+00** |
| | Worst | 2.77E-193 | 2.44E+04 | 8.28E-12 | 3.29E+03 | 1.65E-69 | 1.68E-89 | 8.90E+04 | 2.64E+02 | 5.31E+03 | 2.73E+03 | 6.84E+03 | 2.72E+00 | 0.00E+00 | **0.00E+00** |
| | Mean | 1.35E-194 | 1.38E+04 | 1.89E-12 | 2.43E+03 | 5.62E-71 | 1.30E-90 | 5.84E+04 | 2.17E+02 | 3.50E+03 | 1.51E+03 | 4.25E+03 | 1.01E+00 | 0.00E+00 | **0.00E+00** |
| | STD | 0.00E+00 | 3.24E+03 | 1.69E-12 | 4.00E+02 | 3.02E-70 | 3.09E-90 | 1.30E+04 | 2.26E+01 | 7.32E+02 | 4.62E+02 | 8.52E+02 | 6.69E-01 | 0.00E+00 | **0.00E+00** |
| F2 | Best | 1.75E-139 | 6.96E+01 | 1.31E-08 | 1.72E+01 | 2.78E-56 | 2.55E-47 | 1.95E+02 | 9.57E+00 | 3.40E+01 | 2.94E+01 | 1.26E+01 | 4.08E+00 | 0.00E+00 | **7.75E-28** |
| | Worst | 4.65E-103 | 1.06E+02 | 5.88E-08 | 5.20E+01 | 1.02E-48 | 1.48E-45 | 3.39E+02 | 1.26E+01 | 8.40E+01 | 1.25E+02 | 3.38E+01 | 1.01E+01 | 0.00E+00 | **6.90E-26** |
| | Mean | 1.66E-104 | 8.92E+01 | 3.85E-08 | 2.65E+01 | 6.24E-50 | 4.27E-46 | 2.42E+02 | 1.01E+01 | 6.26E+01 | 5.13E+01 | 1.84E+01 | 6.53E+00 | 0.00E+00 | **1.33E-26** |
| | SID | 8.49E-104 | 9.79E+00 | 1.07E-08 | 7.18E+00 | 2.02E-49 | 4.36E-46 | 2.80E+01 | 7.06E-01 | 1.20E+01 | 1.63E+01 | 4.78E+00 | 1.48E+00 | 0.00E+00 | **2.13E-26** |
| F3 | Best | 2.06E-208 | 2.98E+04 | 3.91E+01 | 1.54E+05 | 6.46E+05 | 5.39E-14 | 1.17E+05 | 3.90E+04 | 3.04E+05 | 1.65E+04 | 8.84E+03 | 2.90E+03 | 0.00E+00 | **6.76E-23** |
| | Worst | 3.71E-132 | 1.86E+05 | 2.93E+03 | 2.33E+05 | 1.71E+06 | 1.16E-08 | 3.35E+05 | 7.08E+04 | 5.73E+05 | 1.01 e+05 | 2.64E+04 | 9.60E+03 | 0.00E+00 | **1.00E-20** |
| | Mean | 1.28E-133 | 9.60E+04 | 6.03E+02 | 2.14E+05 | 1.06E+06 | 7.90E-10 | 2.34E+05 | 5.01E+04 | 4.76E+05 | 5.10E+04 | 1.54E+04 | 4.76E+03 | 0.00E+00 | **1.71E-21** |
| | STD | 6.77E-133 | 3.64E+04 | 6.81E+02 | 1.56E+04 | 2.92E+05 | 2.51E-09 | 6.22E+04 | 8.35E+03 | 6.00E+04 | 2.45E+04 | 4.59E+03 | 1.30E+03 | 0.00E+00 | **2.97E-21** |
| F4 | Best | 2.62E-129 | 2.41E+01 | 1.49E-01 | 8.88E+01 | 2.74E+01 | 7.36E-39 | 8.77E+01 | 1.54E+01 | 9.02E+01 | 2.30E+01 | 1.63E+01 | 8.64E+00 | 0.00E+00 | **8.47E-26** |
| | Worst | 2.74E-100 | 4.00E+01 | 2.37E+00 | 9.76E+01 | 9.77E+01 | 4.44E-37 | 9.74E+01 | 2.75E+01 | 9.80E+01 | 3.75E+01 | 2.32E+01 | 1.37E+01 | 0.00E+00 | **5.72E-23** |
| | Mean | 9.17E-102 | 3.07E+01 | 5.79E-01 | 9.53E+01 | 8.22E+01 | 1.17e-.37 | 9.36E+01 | 2.04E+01 | 9.49E+01 | 2.92E+01 | 1.95E+01 | 1.06E+01 | 0.00E+00 | **6.82E-24** |
| | STD | 5.00E-101 | 3.75E+00 | 4.84E-01 | 1.81E+00 | 1.67E+01 | 1.09E-37 | 2.11E+00 | 2.93E+00 | 1.85E+00 | 3.46E+00 | 1.90E+00 | 1.30E+00 | 0.00E+00 | **1.77E-23** |
| F5 | Best | 4.67E-04 | 2.44E+06 | 9.59E+01 | 4.28E+06 | 9.77E+01 | 9.68E+01 | 6.30E+07 | 4.19E+03 | 2.42E+06 | 4.71E+04 | 3.66E+04 | 5.28E+03 | 2.89E+01 | **9.87E+01** |
| | Worst | 2.88E-01 | 7.32E+06 | 9.81E+01 | 2.10E+07 | 9.81E+01 | 9.84E+01 | 2.56E+08 | 7.36E+03 | 8.88E+06 | 2.55E+05 | 2.63E+05 | 2.59E+04 | 2.90E+01 | **9.89E+01** |
| | Mean | 5.12E-02 | 4.75E+06 | 9.77E+01 | 1.17E+07 | 9.79E+01 | 9.76E+01 | 1.50E+08 | 5.31E+03 | 5.34E+06 | 1.31E+05 | 1.15E+05 | 1.07E+04 | 2.90E+01 | **9.89E+01** |
| | STD | 7.27E-02 | 1.34E+06 | 5.87E-01 | 3.66E+06 | 2.15E-01 | 3.78E-01 | 5.84E+07 | 9.20E+02 | 1.61E+06 | 5.60E+04 | 6.87E+04 | 3.85E+03 | 1.64E-02 | **5.73E-02** |
| F6 | Best | 3.29E-06 | 1.04E+04 | 6.88E+00 | 1.36E+03 | 2.10E+00 | 4.72E+00 | 2.79E+04 | 1.76E+02 | 2.38E+03 | 6.53E+02 | 2.96E+03 | 3.40E+01 | 0.00E+00 | **0.00E+00** |
| | Worst | 3.51E-03 | 1.89E+04 | 1.32E+01 | 3.29E+03 | 6.92E+00 | 9.46E+00 | 1.00E+05 | 2.99E+02 | 4.81E+03 | 2.72E+03 | 8.75E+03 | 1.40E+03 | 0.00E+00 | **0.00E+00** |
| | Mean | 7.23E-04 | 1.44E+04 | 1.00E+01 | 2.41E+03 | 4.26E+00 | 7.39E+00 | 5.91E+04 | 2.30E+02 | 3.39E+03 | 1.52E+03 | 4.65E+03 | 1.27E+02 | 0.00E+00 | **0.00E+00** |
| | STD | 1.00E-03 | 2.38E+03 | 1.55E+00 | 4.63E+02 | 1.25E+00 | 1.08E+00 | 1.45E+04 | 2.57E+01 | 6.31E+02 | 4.43E+02 | 1.10E+03 | 2.50E+02 | 0.00E+00 | **0.00E+00** |
| F7 | Best | 9.38E-06 | 3.15E+00 | 2.54E-03 | 9.39E+00 | 5.81E-05 | 5.17E-04 | 5.24E+01 | 8.27E-02 | 3.39E+00 | 1.50E+00 | 2.15E+00 | 9.33E-01 | 9.29E-05 | **2.38E-90** |
| | Worst | 4.68E-04 | 1.34E+01 | 1.08E-02 | 2.06E+01 | 1.27E-02 | 3.20E-03 | 5.24E+02 | 1.68E-01 | 1.21E+01 | 4.72E+00 | 9.99E+00 | 3.60E+01 | 2.53E-02 | **9.08E-82** |
| | Mean | 1.83E-04 | 7.55E+00 | 6.72E-03 | 1.37E+01 | 3.53E-03 | 1.77E-03 | 2.41E+02 | 1.25E-01 | 6.56E+00 | 2.75E+00 | 4.38E+00 | 4.49E+00 | 6.30E-03 | **1.03E-82** |

| No. | | AVOA | PSO | GWO | FFA | WOA | TLBO | MFO | BBO | DE | SSA | GSA | IPO | SHMS | **FWSC** |
|---|---|---|---|---|---|---|---|---|---|---|---|---|---|---|---|
| | STD | 1.38E-04 | 2.64E+00 | 2.26E-03 | 3.33E+00 | 3.52E-03 | 6.45E-04 | 1.11E+02 | 2.63E-02 | 2.14E+00 | 7.48E-01 | 1.82E+00 | 7.89E+00 | 5.78E-03 | **2.85E-82** |
| F8 | Best | -4.15E+04 | -1.06E+04 | -2.01 e+04 | -1.92E+04 | -4.18E+04 | -2.34E+04 | -2.58E+04 | -2.55E+04 | -1.37E+04 | -2.59E+04 | -6.02E+03 | -1.36E+04 | -4.84E+03 | **0.00E+00** |
| | Worst | -4.12e-F04 | -6.28E+03 | -5.85E+03 | 8.89E+03 | -2.52E+04 | -1.05E+04 | -1.82E+04 | -2.02E+04 | -1.09E+04 | -1.79E+04 | -2.87E+03 | -4.91E+03 | -2.24E+01 | **3.82E+04** |
| | Mean | -4.14E+04 | -7.62E+03 | -1.58E+04 | -1.36E+04 | -3.40E+04 | -1.69E+04 | -2.17E+04 | -2.25E+04 | -1.18E+04 | -2.14E+04 | -4.05E+03 | -1.07E+04 | -1.23E+03 | **3.38E+04** |
| | STD | 5.28E+01 | 1.13E+03 | 3.05E+03 | 3.09E+03 | 5.69E+03 | 2.71E+03 | 1.83E+03 | 1.03E+03 | 6.65E+02 | 1.73E+03 | 7.99E+02 | 1.93E+03 | 1.09E+03 | **1.19E+04** |
| F9 | Best | 0.00E+00 | 6.66E+02 | 1.38E-10 | 6.44E+02 | 0.00E+00 | 0.00E+00 | 7.43E+02 | 2.55E+02 | 9.10E+02 | 1.54E+02 | 1.37E+02 | 2.40E+02 | 0.00E+00 | **0.00E+00** |
| | Worst | 0.00E+00 | 8.35E+02 | 2.39E+01 | 1.08E+03 | 0.00E+00 | 0.00E+00 | 1.00E+03 | 3.98E+02 | 1.02E+03 | 3.03E+02 | 2.65E+02 | 3.43E+02 | 0.00E+00 | **0.00E+00** |
| | Mean | 0.00E+00 | 7.36E+02 | 8.83E+00 | 8.92E+02 | 0.00E+00 | 0.00E+00 | 8.64E+02 | 3.17E+02 | 9.80E+02 | 2.35E+02 | 1.93E+02 | 2.79E+02 | 0.00E+00 | **0.00E+00** |
| | STD | 0.00E+00 | 4.13E+01 | 6.15E+00 | 1.19E+02 | 0.00E+00 | 0.00E+00 | 7.54E+01 | 3.33E+01 | 2.85E+01 | 3.87E+01 | 3.64E+01 | 2.58E+01 | 0.00E+00 | **0.00E+00** |
| F10 | Best | 8.88E-16 | 1.09E+01 | 7.13E-08 | 7.59E+00 | 8.88E-16 | 4.44E-15 | 1.94E+01 | 3.18E+00 | 7.87E+00 | 8.38E+00 | 3.54E+00 | 4.27E+00 | 4.44E-16 | **4.44E-16** |
| | Worst | 8.88E-16 | 1.32E+01 | 3.21E-07 | 9.84E+00 | 7.99E-15 | 7.99E-15 | 2.03E+01 | 3.56E+00 | 1.00E+01 | 1.23E+01 | 6.83E+00 | 6.45E+00 | 4.44E-16 | **4.44E-16** |
| | Mean | 8.88E-16 | 1.19E+01 | 1.30E-07 | 8.68E+00 | 4.08E-15 | 7.63E-15 | 1.99E+01 | 3.42E+00 | 9.11E+00 | 1.01E+01 | 4.96E+00 | 4.93E+00 | 4.44E-16 | **4.44E-16** |
| | STD | 0.00E+00 | 6.76E-01 | 6.01E-08 | 5.81E-01 | 1.94E-15 | 1.05E-15 | 1.31E-01 | 1.19E-01 | 6.18E-01 | 1.10E+00 | 7.62E-01 | 5.55E-01 | 0.00E+00 | **0.00E+00** |
| F11 | Best | 0.00E+00 | 9.56E+01 | 1.93E-13 | 1.48E+01 | 0.00E+00 | 0.00E+00 | 3.55E+02 | 2.65E+00 | 1.64E+01 | 6.66E+00 | 6.10E+02 | 3.47E-01 | 0.00E+00 | **0.00E+00** |
| | Worst | 0.00E+00 | 1.77E+02 | 3.18E-02 | 3.22E+01 | 0.00E+00 | 0.00E+00 | 7.09E+02 | 3.59E+00 | 5.61E+01 | 2.20E+01 | 7.86E+02 | 1.09E+00 | 0.00E+00 | **0.00E+00** |
| | Mean | 0.00E+00 | 1.23E+02 | 5.30E-03 | 2.29E+01 | 0.00E+00 | 0.00E+00 | 5.33E+02 | 3.17E+00 | 3.13E+01 | 1.34E+01 | 6.92E+02 | 8.22E-01 | 0.00E+00 | **0.00E+00** |
| | STD | 0.00E+00 | 2.15E+01 | 1.09E-02 | 4.15E+00 | 0.00E+00 | 0.00E+00 | 1.05E+02 | 2.01E-01 | 8.05E+00 | 3.60E+00 | 4.12E+01 | 2.48E-01 | 0.00E+00 | **0.00E+00** |
| F12 | Best | 4.01E-07 | 4.25E+02 | 1.46E-01 | 1.09E+07 | 1.83E-02 | 7.43E-02 | 5.11E+07 | 1.22E+00 | 3.46E+06 | 1.20E+01 | 4.70E+00 | 5.24E+00 | 0.00E+00 | **9.97E-01** |
| | Worst | 6.36E-05 | 6.15E+05 | 5.08E-01 | 5.79E+07 | 1.30E-01 | 1.62E-01 | 6.24E+08 | 1.27E+01 | 1.94E+07 | 5.70E+01 | 1.93E+01 | 8.15E+00 | 0.00E+00 | **1.17E+00** |
| | Mean | 6.55E-06 | 1.21E+05 | 3.07E-01 | 2.26E+07 | 5.34E-02 | 1.17E-01 | 2.84E+08 | 4.05E+00 | 9.18E+06 | 3.23E+01 | 1.17E+01 | 6.92E+00 | 0.00E+00 | **1.10E+00** |
| | STD | 1.25E-05 | 1.30E+05 | 7.64E-02 | 1.01E+07 | 2.89E-02 | 2.31E-02 | 1.62E+08 | 2.41E+00 | 4.31E+06 | 9.77E+00 | 3.91E+00 | 8.03E-01 | 0.00E+00 | **6.40E-02** |
| F13 | Best | 3.39E-05 | 7.78E+05 | 5.76E+00 | 1.74E+07 | 1.26E+00 | 6.62E+00 | 1.76E+08 | 9.06E+00 | 6.15E+06 | 1.60E+02 | 1.86E+02 | 5.39E+00 | 9.99E+00 | **9.56E+00** |
| | Worst | 2.88E-03 | 8.58E+06 | 7.51E+00 | 8.77E+07 | 4.62E+00 | 9.92E+00 | 1.14E+09 | 1.34E+01 | 3.56E+07 | 5.85E+04 | 3.02E+04 | 1.60E+02 | 1.00E+01 | **9.80E+00** |
| | Mean | 7.54E-04 | 3.39E+06 | 6.72E+00 | 5.33E+07 | 2.75E+00 | 8.09E+00 | 5.71E+08 | 1.13E+01 | 1.68E+07 | 6.04E+03 | 4.90E+03 | 3.59E+01 | 1.00E+01 | **9.65E+00** |
| | STD | 7.67E-04 | 2.04E+06 | 3.97E-01 | 1.87E+07 | 9.29E-01 | 9.62E-01 | 2.90E+08 | 1.18E+00 | 7.34E+06 | 1.19E+04 | 7.72E+03 | 4.23E+01 | 2.91E-03 | **8.19E-02** |

**Table 6 Results of benchmark functions (F1-F13), with 500 dimensions**

| No. | | AVOA | PSO | GWO | FFA | WOA | TLBO | MFO | BBO | DE | SSA | GSA | IPO | SHMS | **FWSC** |
|---|---|---|---|---|---|---|---|---|---|---|---|---|---|---|---|
| F1 | Best | 3.18E-249 | 7.46E+04 | 6.20E-04 | 4.86E+05 | 1.40E-83 | 3.47E-88 | 1.05E+06 | 6.32E+03 | 4.89E+05 | 7.64E+04 | 5.02E+04 | 9.28E+03 | 0.00E+00 | **1.87E-32** |
| | Worst | 2.94E-199 | 1.19E+05 | 2.89E-03 | 5.62E+05 | 6.09E-66 | 4.43E-85 | 1.22E+06 | 8.16E+03 | 6.17E+05 | 1.05E+05 | 6.22E+04 | 1.39E+04 | 0.00E+00 | **1.10E-30** |
| | Mean | 1.04E-200 | 9.80E+04 | 1.57E-03 | 5.25E+05 | 2.05E-67 | 4.13E-86 | 1.12E+06 | 7.15E+03 | 5.63E+05 | 9.33E+04 | 5.57E+04 | 1.13E+04 | 0.00E+00 | **1.81E-31** |
| | STD | 0.00E+00 | 1.30E+04 | 5.47E-04 | 2.02E+04 | 1.11,66 | 9.06E-86 | 3.11E+04 | 5.43E+02 | 2.92E+04 | 8.25E+03 | 2.82E+03 | 1.17E+03 | 0.00E+00 | **3.30E-31** |
| F2 | Best | 2.15E-131 | 4.49E+02 | 8.12E-03 | 1.77E+52 | 2.13E-55 | 4.25E-45 | 5.79E+75 | 2.05E+02 | 1.44E+03 | 4.97E+02 | 2.80E+02 | 1.64E+02 | 0.00E+00 | **2.21E-18** |
| | Worst | 1.00E-99 | 6.36E+02 | 1.42E-02 | 7.09E+117 | 9.69E-47 | 3.44E-43 | 1.77E+120 | 2.75E+02 | 1.54E+03 | 5.75E+02 | 3.23E+270 | 1.98E+02 | 0.00E+00 | **1.19E-17** |
| | Mean | 3.38E-101 | 5.33E+02 | 1.09E-02 | 2.36E+116 | 3.89E-48 | 3.95E-44 | 6.04E+118 | 2.29E+02 | 1.50E+03 | 5.31E+02 | 1.08E+269 | 1.81E+02 | 0.00E+00 | **5.88E-18** |
| | SID | 1.84E-100 | 4.81E+01 | 1.58E-03 | 1.29E+117 | 1.78E-47 | 6.78E-44 | 3.24E+119 | 1.72E+01 | 3.50E+01 | 2.00E+01 | 1.00E+300 | 9.43E+00 | 0.00E+00 | **3.72E-18** |
| F3 | Best | 7.15E-196 | 1.00E+06 | 2.33E+05 | 4.74E+06 | 1.49E+07 | 6.47E-08 | 3.31E+06 | 1.19E+06 | 8.87E+06 | 5.20E+05 | 3.92E+05 | 1.10E+05 | 0.00E+00 | **2.91E-15** |
| | Worst | 8.95E-102 | 5.44E+06 | 5.27E+05 | 6.41E+06 | 5.22E+07 | 8.04E-03 | 6.39E+06 | 2.34E+06 | 1.32E+07 | 2.39E+06 | 3.88E+06 | 2.20E+05 | 0.00E+00 | **1.63E-14** |
| | Mean | 2.98E-103 | 2.29E+06 | 3.57E+05 | 5.57E+06 | 3.04E+07 | 6.76E-04 | 4.80E+06 | 1.55E+06 | 1.16E+07 | 1.24E+06 | 1.17E+06 | 1.48E+05 | 0.00E+00 | **7.69E-15** |
| | STD | 1.63E-102 | 8.31E+05 | 7.70E+04 | 4.80E+05 | 1.00E+07 | 1.99E-03 | 9.16E+05 | 2.25E+05 | 1.10E+06 | 6.10E+05 | 6.80E+05 | 2.78E+04 | 0.00E+00 | **4.77E-15** |
| F4 | Best | 1.87E-136 | 3.44E+01 | 5.69E+01 | 9.83E+01 | 1.62E+01 | 8.57E-37 | 9.82E+01 | 4.86E+01 | 9.88E+01 | 3.50E+01 | 2.55E+01 | 1.84E+01 | 0.00E+00 | **1.04E-17** |
| | Worst | 3.49E-100 | 4.77E+01 | 7.77E+01 | 9.92E+01 | 9.87E+01 | 4.51E-35 | 9.91E+01 | 5.75E+01 | 9.92E+01 | 4.86E+01 | 3.27E+01 | 2.19E+01 | 0.00E+00 | **2.66E-16** |
| | Mean | 1.47E-101 | 3.82E+01 | 6.47E+01 | 9.89E+01 | 8.19E+01 | 8.65e-.36 | 9.85E+01 | 5.27E+01 | 9.89E+01 | 4.03E+01 | 2.87E+01 | 2.04E+01 | 0.00E+00 | **7.68E-17** |
| | STD | 6.43E-101 | 2.73E+00 | 5.77E+00 | 3.01E-01 | 2.10E+01 | 9.83E-36 | 3.71E-01 | 2.07E+00 | 1.98E-01 | 3.30E+00 | 1.56E+00 | 9.22E-01 | 0.00E+00 | **7.84E-17** |
| F5 | Best | 3.34E-03 | 2.16E+07 | 4.93E+02 | 1.72E+09 | 4.91E+02 | 4.94E+02 | 4.64E+09 | 6.08E+05 | 1.75E+09 | 2.75E+07 | 6.49E+06 | 2.31E+06 | 2.89E+01 | **4.99E+02** |
| | Worst | 1.86E+01 | 6.50E+07 | 4.96E+02 | 2.95E+09 | 4.97E+02 | 4.97E+02 | 5.45E+09 | 1.01E+06 | 6.00E+09 | 4.72E+07 | 1.40E+07 | 3.53E+06 | 2.90E+01 | **4.99E+02** |
| | Mean | 3.66E+00 | 4.31E+07 | 4.94E+02 | 2.30E+09 | 4.93E+02 | 4.95E+02 | 5.02E+09 | 8.44E+05 | 2.82E+09 | 3.71E+07 | 8.69E+06 | 2.85E+06 | 2.90E+01 | **4.99E+02** |
| | STD | 4.12E+00 | 1.13E+07 | 3.30E-01 | 3.17E+08 | 3.52E-01 | 1.13E-01 | 2.21E+08 | 9.37E+04 | 8.04E+08 | 4.14E+06 | 1.62E+06 | 3.81E+05 | 1.83E-02 | **5.44E-02** |
| F6 | Best | 2.76E-04 | 7.35E+04 | 8.75E+01 | 4.72E+05 | 1.89E+01 | 9.06E+01 | 1.05E+06 | 6.66E+03 | 4.87E+05 | 7.57E+04 | 5.00E+04 | 1.36E+04 | 0.00E+00 | **0.00E+00** |
| | Worst | 4.78E-01 | 1.20E+05 | 9.44E+01 | 5.73E+05 | 4.73E+01 | 9.82E+01 | 1.21E+06 | 8.51E+03 | 6.23E+05 | 1.08E+05 | 6.27E+04 | 2.71E+04 | 0.00E+00 | **0.00E+00** |
| | Mean | 5.90E-02 | 9.65E+04 | 9.14E+01 | 5.29E+05 | 3.24E+01 | 9.42E+01 | 1.15E+06 | 7.35E+03 | 5.50E+05 | 9.42E+04 | 5.70E+04 | 1.96E+04 | 0.00E+00 | **0.00E+00** |
| | STD | 1.03E-01 | 1.42E+04 | 1.68E+00 | 2.55E+04 | 8.30E+00 | 2.05E+00 | 3.50E+04 | 4.77E+02 | 3.53E+04 | 6.51E+03 | 2.75E+03 | 4.06E+03 | 0.00E+00 | **0.00E+00** |
| F7 | Best | 2.71E-06 | 1.74E+02 | 2.62E-02 | 1.23E+04 | 1.43E-04 | 5.14E-04 | 3.28E+04 | 3.67E+02 | 1.16E+04 | 2.08E+02 | 7.08E+02 | 1.77E+03 | 2.60E-03 | **6.48E-60** |
| | Worst | 5.53E-04 | 6.70E+02 | 7.89E-02 | 1.91E+04 | 1.40E-02 | 3.30E-03 | 4.23E+04 | 7.03E+02 | 2.04E+04 | 3.54E+02 | 1.44E+03 | 4.79E+03 | 2.84E-02 | **1.29E-55** |
| | Mean | 2.09E-04 | 3.50E+02 | 5.20E-02 | 1.59E+04 | 4.65E-03 | 1.67E-03 | 3.87E+04 | 4.96E+02 | 1.55E+04 | 2.75E+02 | 9.88E+02 | 2.86E+03 | 1.02E-02 | **2.38E-56** |
| | STD | 1.54E-04 | 1.35E+02 | 1.36E-02 | 1.49E+03 | 4.70E-03 | 6.18E-04 | 2.31E+03 | 9.00E+01 | 2.19E+03 | 3.96E+01 | 1.74E+02 | 7.30E+02 | 6.91E-03 | **3.92E-56** |

| No. | | AVOA | PSO | GWO | FFA | WOA | TLBO | MFO | BBO | DE | SSA | GSA | IPO | SHMS | **FWSC** |
|---|---|---|---|---|---|---|---|---|---|---|---|---|---|---|---|
| F8 | Best | -2.13e-F05 | -2.25E+04 | -6.50E+04 | -3.92E+04 | -2.09E+05 | -6.07E+04 | -7.14E+04 | -7.49E+04 | -2.76E+04 | -6.80E+04 | -1.61E+04 | -4.85E+04 | -7.90E+03 | **1.97E+05** |
| | Worst | -2.10e-F05 | -1.34E+04 | -4.96E+04 | -1.84E+04 | -1.26E+05 | -2.57E+04 | -5.01E+04 | -6.62E+04 | -2.38E+04 | -4.93E+04 | -6.32E+03 | -9.29E+03 | -5.07E+01 | **2.02E+05** |
| | Mean | -2.12E+05 | -1.78E+04 | -5.80E+04 | -2.73E+04 | -1.69E+05 | -3.96E+04 | -6.22E+04 | -7.07E+04 | -2.55E+04 | -6.08E+04 | -1.10E+04 | -3.06E+04 | -2.75E+03 | **2.00E+05** |
| | STD | 6.16E+02 | 2.57E+03 | 3.75E+03 | 5.92E+03 | 3.09E+04 | 9.39E+03 | 4.89E+03 | 2.53E+03 | 1.12E+03 | 4.37E+03 | 2.27E+03 | 1.44E+04 | 2.76E+03 | **1.49E+03** |
| F9 | Best | 0.00e-F00 | 4.28E+03 | 2.37E+01 | 6.50E+03 | 0.00E+00 | 0.00E+00 | 6.61E+03 | 5.75E+03 | 6.51E+03 | 2.95E+03 | 2.53E+03 | 3.17E+03 | 0.00E+00 | **0.00E+00** |
| | Worst | 0.00e-F00 | 4.86E+03 | 1.61E+02 | 7.02E+03 | 0.00E+00 | 0.00E+00 | 7.27E+03 | 6.41E+03 | 7.10E+03 | 3.34E+03 | 2.98E+03 | 3.45E+03 | 0.00E+00 | **0.00E+00** |
| | Mean | 0.00E+00 | 4.61E+03 | 7.88E+01 | 6.84E+03 | 0.00E+00 | 0.00E+00 | 6.93E+03 | 6.05E+03 | 6.75E+03 | 3.16E+03 | 2.73E+03 | 3.33E+03 | 0.00E+00 | **0.00E+00** |
| | STD | 0.00e-F00 | 1.58E+02 | 2.83E+01 | 1.16E+02 | 0.00E+00 | 0.00E+00 | 1.65E+02 | 1.80E+02 | 1.24E+02 | 1.04E+02 | 1.16E+02 | 7.02E+01 | 0.00E+00 | **0.00E+00** |
| F10 | Best | 8.88E-16 | 1.21E+01 | 1.23E-03 | 1.95E+01 | 8.88E-16 | 4.44E-15 | 2.03E+01 | 2.02E+01 | 1.91E+01 | 1.38E+01 | 1.01E+01 | 1.35E+01 | 4.44E-16 | **4.44E-16** |
| | Worst | 8.88E-16 | 1.44E+01 | 3.38E-03 | 1.98E+01 | 7.99E-15 | 7.99E-15 | 2.05E+01 | 2.05E+01 | 2.01E+01 | 1.46E+01 | 1.11E+01 | 1.46E+01 | 4.44E-16 | **4.44E-16** |
| | Mean | 8.88E-16 | 1.30E+01 | 1.90E-03 | 1.97E+01 | 4.44E-15 | 7.87E-15 | 2.04E+01 | 2.03E+01 | 1.95E+01 | 1.42E+01 | 1.05E+01 | 1.41E+01 | 4.44E-16 | **4.44E-16** |
| | STD | 0.00e-F00 | 6.47E-01 | 4.40E-04 | 9.58E-02 | 2.46E-15 | 6.48E-16 | 1.49E-01 | 8.64E-02 | 1.45E-01 | 2.27E-01 | 2.32E-01 | 2.56E-01 | 0.00E+00 | **0.00E+00** |
| F11 | Best | 0.00E+00 | 6.73E+02 | 1.07E-04 | 4.18E+03 | 0.00E+00 | 0.00E+00 | 9.83E+03 | 2.27E+03 | 4.02E+03 | 7.64E+02 | 8.17E+03 | 8.63E+01 | 0.00E+00 | **0.00E+00** |
| | Worst | 0.00e-F00 | 1.28E+03 | 1.29E-01 | 5.16E+03 | 0.00E+00 | 0.00E+00 | 1.06E+04 | 3.30E+03 | 5.53E+03 | 9.47E+02 | 8.99E+03 | 1.11E+02 | 0.00E+00 | **0.00E+00** |
| | Mean | 0.00e-F00 | 9.41E+02 | 3.35E-02 | 4.71E+03 | 0.00E+00 | 0.00E+00 | 1.02E+04 | 3.02E+03 | 5.03E+03 | 8.46E+02 | 8.618+03 | 9.618+01 | 0.00E+00 | **0.00E+00** |
| | STD | 0.00E+00 | 1.74E+02 | 4.90E-02 | 2.14E+02 | 0.00E+00 | 0.00E+00 | 2.86E+02 | 2.62E+02 | 3.34E+02 | 5.56E+01 | 2.05E+02 | 5.73E+00 | 0.00E+00 | **0.00E+00** |
| F12 | Best | 8.97E-09 | 8.55E+05 | 6.59E-01 | 4.04E+09 | 2.45E-02 | 5.89E-01 | 1.05E+10 | 3.84E+08 | 4.69E+09 | 2.27E+05 | 1.99E+02 | 1.39E+01 | 0.00E+00 | **1.13E+00** |
| | Worst | 4.98E-04 | 1.47E+07 | 8.21E-01 | 9.90E+09 | 1.87E-01 | 7.19E-01 | 1.24E+10 | 7.24E+08 | 1.83E+10 | 3.69E+06 | 8.02E+04 | 2.42E+01 | 0.00E+00 | **1.17E+00** |
| | Mean | 4.19E-05 | 4.05E+06 | 7.43E-01 | 7.44E+09 | 8.59E-02 | 6.52E-01 | 1.20E+10 | 4.95E+08 | 1.13E+10 | 1.39E+06 | 1.46E+04 | 1.92E+01 | 0.00E+00 | **1.15E+00** |
| | STD | 9.45E-05 | 3.49E+06 | 4.36E-02 | 1.53E+09 | 4.24E-02 | 3.33E-02 | 5.00E+08 | 8.51E+07 | 4.24E+09 | 7.92E+05 | 1.91E+04 | 2.40E+00 | 0.00E+00 | **1.11E-02** |
| F13 | Best | 3.38E-04 | 1.41E+07 | 4.88E+01 | 7.74E+09 | 9.45E+00 | 4.98E+01 | 1.94E+10 | 8.71E+08 | 7.35E+09 | 1.85E+07 | 1.69E+06 | 6.78E+03 | 5.00E+01 | **4.96E+01** |
| | Worst | 1.99E-01 | 1.27E+08 | 5.40E+01 | 1.42E+10 | 2.58E+01 | 4.99E+01 | 2.42E+10 | 1.75E+09 | 3.43E+10 | 5.92E+07 | 7.18E+06 | 5.47E+04 | 5.00E+01 | **4.97E+01** |
| | Mean | 2.52E-02 | 5.34E+07 | 5.11E+01 | 1.13E+10 | 1.80E+01 | 4.99E+01 | 2.21E+10 | 1.35E+09 | 1.47E+10 | 3.32E+07 | 3.89E+06 | 2.73E+04 | 5.00E+01 | **4.97E+01** |
| | STD | 4.30E-02 | 2.83E+07 | 1.25E+00 | 1.63E+09 | 4.12E+00 | 7.42E-03 | 9.85E+08 | 2.47E+08 | 6.31E+09 | 8.49E+06 | 1.33E+08 | 1.13E+04 | 2.41E-03 | **4.71E-02** |

Table 7 Results of benchmark functions (F1-F13), with 1000 dimensions

| No. | | AVOA | PSO | GWO | FFA | WOA | TLBO | MFO | BBO | DE | SSA | GSA | IPO | SHMS | **FWSC** |
|---|---|---|---|---|---|---|---|---|---|---|---|---|---|---|---|
| F1 | Best | 7.50E-278 | 1.47E+05 | 1.41E-01 | 1.33E+06 | 9.06E-84 | 2.06E-87 | 2.62E+06 | 6.00E+05 | 1.49E+06 | 2.12E+05 | 1.22E+05 | 4.10E+04 | 0.00E+00 | **1.06E-27** |
| | Worst | 3.06E-193 | 3.04E+05 | 4.79E-01 | 1.49E+06 | 5.41E-67 | 2.33E-84 | 2.83E+06 | 7.30E+05 | 1.81E+06 | 2.67E+05 | 1.48E+05 | 5.07E+04 | 0.00E+00 | **6.02E-26** |
| | Mean | 1.05E-194 | 2.16E+05 | 2.42E-01 | 1.43E+06 | 1.80E-68 | 2.44E-85 | 2.73E+06 | 6.69E+05 | 1.60E+06 | 2.37E+05 | 1.31E+05 | 4.60E+04 | 0.00E+00 | **1.23E-26** |
| | STD | 0.00E+00 | 4.33E+04 | 7.43E-02 | 4.49E+04 | 9.88E-68 | 4.40E-85 | 4.83E+04 | 2.99E+04 | 8.51E+04 | 1.30E+04 | 5.62E+03 | 2.32E+03 | 0.00E+00 | **1.77E-26** |
| F2 | Best | 4.21E-173 | 1.00E+300 | 2.87E-01 | 1.00E+300 | 2.13E-56 | 1.00E+300 | 1.00E+300 | 1.00E+300 | 1.00E+300 | 1.11E+03 | 1.17E+263 | 4.39E+02 | 0.00E+00 | **7.50E-16** |
| | Worst | 2.02E-112 | 1.00E+300 | 2.10E+00 | 1.00E+300 | 2.83E-47 | 1.00E+300 | 1.00E+300 | 1.00E+300 | 1.00E+300 | 1.25E+03 | 6.38E+289 | 4.88E+02 | 0.00E+00 | **2.79E-15** |
| | Mean | 6.75E-114 | 1.00E+300 | 7.17E-01 | 1.00E+300 | 1.93E-48 | 1.00E+300 | 1.00E+300 | 1.00E+300 | 1.00E+300 | 1.19E+03 | 3.46E+288 | 4.60E+02 | 0.00E+00 | **1.87E-15** |
| | SID | 3.69E-113 | 1.00E+300 | 3.96E-01 | 1.00E+300 | 5.68E-48 | 1.00E+300 | 1.00E+300 | 1.00E+300 | 1.00E+300 | 3.14E+01 | 1.00E+300 | 1.04E+01 | 0.00E+00 | **6.32E-16** |
| F3 | Best | 2.27E-219 | 3.96E+06 | 1.15E+06 | 1.83E+07 | 6.83E+07 | 4.55E-06 | 1.31E+07 | 6.81E+06 | 3.62E+07 | 2.37E+06 | 2.99E+06 | 3.28E+05 | 0.00E+00 | **9.29E-14** |
| | Worst | 5.26E-111 | 1.36E+07 | 2.65E+06 | 2.49E+07 | 2.26E+08 | 1.59E-01 | 2.66E+07 | 1.23E+07 | 5.94E+07 | 9.55E+06 | 1.37E+07 | 7.63E+05 | 0.00E+00 | **2.62E-12** |
| | Mean | 1.76E-112 | 8.18E+06 | 1.67E+06 | 2.15E+07 | 1.32E+08 | 1.13E-02 | 1.85E+07 | 9.66E+06 | 4.77E+07 | 5.92E+06 | 6.53E+06 | 5.55E+05 | 0.00E+00 | **8.51E-13** |
| | STD | 9.60E-112 | 2.44E+06 | 3.22E+05 | 1.82E+06 | 4.84E+07 | 3.16E-02 | 3.66E+06 | 1.60E+06 | 5.67E+06 | 1.92E+06 | 2.56E+06 | 9.61E+04 | 0.00E+00 | **7.78E-13** |
| F4 | Best | 3.10E-143 | 3.61E+01 | 7.28E+01 | 9.91E+01 | 9.05E+00 | 2.61E-36 | 9.90E+01 | 8.27E+01 | 9.88E+01 | 4.00E+01 | 3.13E+01 | 2.24E+01 | 0.00E+00 | **4.33E-15** |
| | Worst | 1.34E-100 | 5.58E+01 | 8.84E+01 | 9.95E+01 | 9.96E+01 | 1.50E-34 | 9.94E+01 | 8.99E+01 | 9.95E+01 | 5.26E+01 | 3.72E+01 | 2.47E+01 | 0.00E+00 | **1.89E-13** |
| | Mean | 4.47E-102 | 4.23E+01 | 7.90E+01 | 9.92E+01 | 8.23E+01 | 2.97E-35 | 9.92E+01 | 8.60E+01 | 9.91E+01 | 4.48E+01 | 3.40E+01 | 2.35E+01 | 0.00E+00 | **2.70E-14** |
| | STD | 2.44E-101 | 3.75E+00 | 3.21E+00 | 9.52E-02 | 2.01E+01 | 2.95E-35 | 1.42E-01 | 2.03E+00 | 1.40E-01 | 2.82E+00 | 1.28E+00 | 7.51E-01 | 0.00E+00 | **5.72E-14** |
| F5 | Best | 1.44E-03 | 5.25E+07 | 1.00E+03 | 6.94E+09 | 9.92E+02 | 9.93E+02 | 1.19E+10 | 7.96E+08 | 1.42E+10 | 9.60E+07 | 2.09E+07 | 1.14E+07 | 2.89E+01 | **9.99E+02** |
| | Worst | 5.25E+01 | 2.02E+08 | 1.12E+03 | 9.62E+09 | 9.95E+02 | 9.95E+02 | 1.31E+10 | 1.17E+09 | 1.53E+10 | 1.53E+08 | 2.85E+07 | 1.61E+07 | 2.90E+01 | **9.99E+02** |
| | Mean | 5.80E+00 | 9.78E+07 | 1.02E+03 | 8.51E+09 | 9.93E+02 | 9.94E+02 | 1.24E+10 | 9.58E+08 | 1.48E+10 | 1.14E+08 | 2.47E+07 | 1.36E+07 | 2.90E+01 | **9.99E+02** |
| | STD | 1.11E+01 | 3.42E+07 | 2.12E+01 | 6.35E+08 | 7.07E-01 | 1.31E-01 | 2.91E+08 | 8.91E+07 | 2.77E+08 | 1.21E+07 | 1.95E+06 | 1.09E+06 | 2.18E-02 | **1.26E-02** |
| F6 | Best | 2.26E-03 | 1.41E+05 | 1.95E+02 | 1.28E+06 | 2.80E+01 | 2.03E+02 | 2.57E+06 | 6.07E+05 | 1.38E+06 | 2.15E+05 | 1.21E+05 | 4.69E+04 | 0.00E+00 | **0.00E+00** |
| | Worst | 1.05E+00 | 2.91E+05 | 2.06E+02 | 1.47E+06 | 1.17E+02 | 2.18E+02 | 2.75E+06 | 7.21E+05 | 2.27E+06 | 2.54E+05 | 1.48E+05 | 9.84E+04 | 0.00E+00 | **0.00E+00** |
| | Mean | 1.27E-01 | 1.99E+05 | 2.00E+02 | 1.41E+06 | 7.29E+01 | 2.13E+02 | 2.72E+06 | 6.66E+05 | 1.63E+06 | 2.37E+05 | 1.31E+05 | 6.24E+04 | 0.00E+00 | **0.00E+00** |
| | STD | 2.06E-01 | 3.44E+04 | 2.79E+00 | 5.50E+04 | 1.91E+01 | 2.20E+00 | 4.95E+04 | 3.34E+04 | 1.61E+05 | 1.12E+04 | 5.65E+03 | 1.06E+04 | 0.00E+00 | **0.00E+00** |
| F7 | Best | 1.41E-05 | 8.87E+02 | 9.98E-02 | 8.40E+04 | 1.30E-04 | 8.77E-04 | 1.90E+05 | 1.19E+04 | 9.92E+04 | 1.31E+03 | 5.43E+03 | 1.83E+04 | 2.60E-03 | **3.80E-50** |
| | Worst | 9.11E-04 | 2.64E+03 | 2.10E-01 | 1.28E+05 | 1.87E-02 | 3.67E-03 | 2.09E+05 | 1.56E+04 | 2.49E+05 | 2.15E+03 | 8.26E+03 | 2.69E+04 | 2.84E-02 | **1.79E-46** |

| No. | | AVOA | PSO | GWO | FFA | WOA | TLBO | MFO | BBO | DE | SSA | GSA | IPO | SHMS | **FWSC** |
|---|---|---|---|---|---|---|---|---|---|---|---|---|---|---|---|
| | Mean | 2.85E-04 | 1.57E+03 | 1.48E-01 | 1.10E+05 | 3.54E-03 | 2.01E-03 | 1.97E+05 | 1.34E+04 | 2.05E+05 | 1.69E+03 | 6.43E+03 | 2.20E+04 | 1.02E-02 | **3.23E-47** |
| | STD | 2.43E-04 | 4.93E+02 | 3.08E-02 | 1.27E+04 | 4.43E-03 | 6.18E-04 | 4.50E+03 | 1.15E+03 | 5.24E+04 | 1.74E+02 | 7.09E+02 | 2.00E+03 | 6.91E-03 | **5.85E-47** |
| F8 | Best | -4.18E+05 | -3.43E+04 | -1.02E+05 | -6.79E+04 | -4.18E+05 | -9.18E+04 | -1.79E+05 | -8.37E+04 | -4.25E+04 | -1.01E+05 | -2.38E+04 | -7.88E+04 | 0.00E+00 | **4.01E+05** |
| | Worst | -3.95E+05 | -1.98E+04 | -1.77E+04 | -2.46E+04 | -2.21E+05 | -2.75E+04 | -7.40E+04 | -7.22E+04 | -3.39E+04 | -7.11E+04 | -9.51E+03 | -1.60E+04 | 0.00E+00 | **4.08E+05** |
| | Mean | -4.17E+05 | -2.58E+04 | -8.58E+04 | -3.72E+04 | -3.27E+05 | -5.87E+04 | -8.72E+04 | -7.80E+04 | -3.64E+04 | -8.72E+04 | -1.34E+04 | -5.47E+04 | 0.00E+00 | **4.05E+05** |
| | STD | 4.69E+03 | 3.72E+03 | 1.88E+04 | 1.07E+04 | 6.28E+04 | 1.38E+04 | 6.85E+03 | 2.85E+03 | 2.00E+03 | 8.08E+03 | 2.78E+03 | 1.89E+04 | 0.00E+00 | **2.02E+03** |
| F9 | Best | 0.00E+00 | 9.21E+03 | 1.11E+02 | 1.38E+04 | 0.00E+00 | 0.00E+00 | 1.50E+04 | 1.13E+04 | 1.38E+04 | 7.17E+03 | 6.23E+03 | 7.62E+03 | 0.00E+00 | **0.00E+00** |
| | Worst | 0.00E+00 | 1.00E+04 | 2.90E+02 | 1.46E+04 | 0.00E+00 | 0.00E+00 | 1.58E+04 | 1.18E+04 | 1.43E+04 | 7.93E+03 | 7.02E+03 | 8.10E+03 | 0.00E+00 | **0.00E+00** |
| | Mean | 0.00E+00 | 9.66E+03 | 1.90E+02 | 1.43E+04 | 0.00E+00 | 0.00E+00 | 1.52E+04 | 1.17E+04 | 1.41E+04 | 7.57E+03 | 6.69E+03 | 7.79E+03 | 0.00E+00 | **0.00E+00** |
| | STD | 0.00E+00 | 1.87E+02 | 4.19E+01 | 1.64E+02 | 0.00E+00 | 0.00E+00 | 1.92E+02 | 2.42E+02 | 2.41E+02 | 2.07E+02 | 2.16E+02 | 1.13E+02 | 0.00E+00 | **0.00E+00** |
| F10 | Best | 8.88E-16 | 1.24E+01 | 1.38E-02 | 1.95E+01 | 8.88E-16 | 7.99E-15 | 2.00E+01 | 1.96E+01 | 2.01E+01 | 1.42E+01 | 1.10E+01 | 1.41E+01 | 4.44E-16 | **4.44E-16** |
| | Worst | 8.88E-16 | 1.55E+01 | 2.24E-02 | 2.03E+01 | 4.44E-15 | 1.30E+01 | 2.06E+01 | 2.03E+01 | 2.06E+01 | 1.49E+01 | 1.15E+01 | 1.48E+01 | 4.44E-16 | **4.00E-15** |
| | Mean | 8.88E-16 | 1.37E+01 | 1.83E-02 | 2.00E+01 | 3.01E-15 | 4.29E-01 | 2.05E+01 | 2.00E+01 | 2.03E+01 | 1.46E+01 | 1.12E+01 | 1.45E+01 | 4.44E-16 | **2.58E-15** |
| | STD | 0.00E+00 | 8.96E-01 | 2.17E-03 | 9.92E-02 | 1.77E-15 | 2.34E+00 | 1.91E-01 | 8.64E-02 | 1.10E-01 | 1.82E-01 | 1.43E-01 | 1.51E-01 | 0.00E+00 | **1.83E-15** |
| F11 | Best | 0.00E+00 | 1.42E+03 | 8.47E-03 | 1.12E+04 | 0.00E+00 | 0.00E+00 | 2.40E+04 | 5.15E+03 | 1.24E+04 | 1.97E+03 | 1.98E+04 | 4.78E+02 | 0.00E+00 | **0.00E+00** |
| | Worst | 0.00E+00 | 2.50E+03 | 2.50E-01 | 1.38E+04 | 0.00E+00 | 1.11E-16 | 2.51E+04 | 6.47E+03 | 1.70E+04 | 2.30E+03 | 2.13E+04 | 5.97E+02 | 0.00E+00 | **0.00E+00** |
| | Mean | 0.00E+00 | 1.87E+03 | 5.11E-02 | 1.28E+04 | 0.00E+00 | 3.33E-17 | 2.46E+04 | 5.86E+03 | 1.45E+04 | 2.07E+03 | 2.05E+04 | 5.44E+02 | 0.00E+00 | **0.00E+00** |
| | STD | 0.00E+00 | 2.73E+02 | 7.61E-02 | 5.92E+02 | 0.00E+00 | 5.17E-17 | 3.53E+02 | 3.55E+02 | 1.02E+03 | 8.45E+01 | 3.34E+02 | 2.69E+01 | 0.00E+00 | **0.00E+00** |
| F12 | Best | 2.19E-07 | 1.28E+06 | 9.04E-01 | 2.38E+10 | 3.60E-02 | 8.20E-01 | 2.79E+10 | 7.51E+08 | 3.39E+10 | 5.05E+06 | 5.39E+04 | 2.68E+01 | 0.00E+00 | **1.16E+00** |
| | Worst | 5.89E-04 | 3.48E+07 | 1.88E+00 | 3.85E+10 | 3.03E-01 | 8.94E-01 | 3.22E+10 | 1.07E+09 | 3.79E+10 | 1.79E+07 | 5.00E+05 | 1.90E+02 | 0.00E+00 | **1.17E+00** |
| | Mean | 6.76E-05 | 1.09E+07 | 1.18E+00 | 3.17E+10 | 1.05E-01 | 8.54E-01 | 3.04E+10 | 8.87E+08 | 3.68E+10 | 1.11E+07 | 1.88E+05 | 5.47E+01 | 0.00E+00 | **1.17E+00** |
| | STD | 1.44E-04 | 8.13E+06 | 2.76E-01 | 4.68E+09 | 6.06E-02 | 1.52E-02 | 1.01E+09 | 8.29E+07 | 1.09E+09 | 3.00E+06 | 1.19E+05 | 3.99E+01 | 0.00E+00 | **3.81E-03** |
| F13 | Best | 1.93E-03 | 3.71E+07 | 1.06E+02 | 3.44E+10 | 1.91E+01 | 9.98E+01 | 5.15E+10 | 2.35E+09 | 6.49E+10 | 1.08E+08 | 9.50E+06 | 3.58E+05 | 1.00E+02 | **9.96E+01** |
| | Worst | 2.89E-01 | 3.20E+08 | 1.24E+02 | 4.87E+10 | 6.70E+01 | 9.95E+01 | 5.80E+10 | 3.42E+09 | 6.92E+10 | 2.15E+08 | 2.66E+07 | 1.04E+06 | 1.00E+02 | **9.98E+01** |
| | Mean | 5.76E-02 | 1.15E+08 | 1.18E+02 | 4.44E+10 | 3.65E+01 | 9.94E+01 | 5.55E+10 | 2.78E+09 | 6.70E+10 | 1.46E+08 | 1.60E+07 | 6.38E+05 | 1.00E+02 | **9.97E+01** |
| | STD | 8.02E-02 | 6.11E+07 | 5.35E+00 | 3.17E+09 | 1.14E+01 | 1.17E-02 | 1.44E+09 | 2.71E+08 | 1.22E+09 | 2.20E+07 | 4.03E+06 | 1.92E+05 | 3.33E-03 | **4.86E-02** |

Table 8 Results of benchmark functions (F14-F23)

| No. | | AVOA | PSO | GWO | FFA | WOA | TLBO | MFO | BBO | DE | SSA | GSA | IPO | SHMS | **FWSC** |
|---|---|---|---|---|---|---|---|---|---|---|---|---|---|---|---|
| F14 | Best | 9.98E-01 | 9.98E-01 | 9.98E-01 | 9.98E-01 | 9.98E-01 | 9.98E-01 | 9.98E-01 | 9.98E-01 | 9.98E-01 | 9.98E-01 | 9.98E-01 | 9.98E-01 | 1.27E+01 | **1.01E+00** |
| | Worst | 2.98E+00 | 1.60E+01 | 1.27E+01 | 7.83B+00 | 1.08E+01 | 9.98E-01 | 1.27E+01 | 7.87E+00 | 5.93E+00 | 4.95E+00 | 1.25E+01 | 6.90E+00 | 1.27E+01 | **1.17E+01** |
| | Mean | 1.26E+00 | 5.95E+00 | 4.06E+00 | 1.91E+00 | 2.57E+00 | 9.98E-01 | 2.87E+00 | 3.27E+00 | 1.39E+00 | 1.39E+00 | 5.41E+00 | 2.61E+00 | 1.27E+01 | **1.17E+01** |
| | STD | 5.79E-01 | 3.76E+00 | 4.18E+00 | 2.07E+00 | 2.54E+00 | 4.12E-17 | 2.55E+00 | 2.12E+00 | 1.26E+00 | 8.85E-01 | 3.35E+00 | 1.82E+00 | 0.00E+00 | **1.64E-10** |
| F15 | Best | 3.08E-04 | 4.20E-04 | 3.07E-04 | 3.34B-04 | 3.08E-04 | 3.07E-04 | 7.12E-04 | 3.93E-04 | 3.08E-04 | 3.08E-04 | 1.17E-03 | 3.07E-04 | 4.67E-04 | **3.35E-04** |
| | Worst | 7.57E-04 | 5.67E-02 | 2.04E-02 | 8.75E-04 | 2.13E-03 | 2.04E-02 | 2.01E-02 | 2.04E-02 | 2.01E-02 | 6.33E-02 | 1.29E-02 | 7.14E-04 | 7.65E-03 | **2.10E-02** |
| | Mean | 4.65E-04 | 1.07E-02 | 4.39E-03 | 5.90E-04 | 6.55E-04 | 1.09E-03 | 1.93E-03 | 4.79E-03 | 1.14E-03 | 4.28E-03 | 4.35E-03 | 4.30E-04 | 2.75E-03 | **1.64E-03** |
| | STD | 1.48E-04 | 1.80E-02 | 8.13E-03 | 1.19B-04 | 4.22E-04 | 3.65E-03 | 3.75E-03 | 7.92E-03 | 3.64E-03 | 1.22E-02 | 2.74E-03 | 1.36E-04 | 1.89E-03 | **1.64E-02** |
| F16 | Best | -1.03E+00 | -1.03E+00 | -1.03E+00 | -1.03E+00 | -1.03E+00 | -1.03E+00 | -1.03E+00 | -1.03E+00 | -1.03E+00 | -1.03E+00 | -1.03E+00 | -1.03E+00 | -1.03E+00 | **-1.03E+00** |
| | Worst | -1.03E+00 | -1.03E+00 | -1.03E+00 | -1.03E+00 | -1.03E+00 | -1.03E+00 | -1.03E+00 | -1.03E+00 | -1.03E+00 | -1.03E+00 | -1.03E+00 | -1.03E+00 | -1.01E+00 | **-9.96E-01** |
| | Mean | -1.03E+00 | -1.03E+00 | -1.03E+00 | -1.03E+00 | -1.03E+00 | -1.03E+00 | -1.03E+00 | -1.03E+00 | -1.03E+00 | -1.03E+00 | -1.03E+00 | -1.03E+00 | -1.03E+00 | **-1.03E+00** |
| | STD | 6.78E-16 | 6.59E-06 | 1.56E-08 | 6.78E-16 | 2.02E-09 | 6.78E-16 | 6.78E-16 | 4.43E-12 | 6.78E-16 | 4.09E-14 | 4.88E-16 | 5.61E-16 | 4.95E-03 | **6.55E-03** |
| F17 | Best | 3.98E-01 | 3.98E-01 | 3.98E-01 | 3.98E-01 | 3.98E-01 | 3.98E-01 | 3.98E-01 | 3.98E-01 | 3.98E-01 | 3.98E-01 | 3.98E-01 | 3.98E-01 | 3.16E+00 | **3.98E-01** |
| | Worst | 3.98E-01 | 8.86E-01 | 3.98E-01 | 3.98E-01 | 3.98E-01 | 3.98E-01 | 3.98E-01 | 8.92E-01 | 3.98E-01 | 3.98E-01 | 3.98E-01 | 3.98E-01 | 1.77E+01 | **4.06E-01** |
| | Mean | 3.98E-01 | 4.62E-01 | 3.98E-01 | 3.98E-01 | 3.98E-01 | 3.98E-01 | 3.98E-01 | 4.77E-01 | 3.98E-01 | 3.98E-01 | 3.98E-01 | 3.98E-01 | 9.72E+00 | **4.00E-01** |
| | STD | 1.69E-16 | 1.15E-01 | 1.47E-06 | 3.95E-12 | 6.06E-06 | 1.69E-16 | 1.69E-16 | 1.39E-01 | 1.69E-16 | 2.29E-05 | 1.69E-16 | 2.87E-13 | 3.31E+00 | **2.28E-03** |
| F18 | Best | 3.00E+00 | 3.00E+00 | 3.00E+00 | 3.00E+00 | 3.00E+00 | 3.00E+00 | 3.00E+00 | 3.00E+00 | 3.00E+00 | 3.00E+00 | 3.00E+00 | 3.00E+00 | 3.22E+00 | **3.00E+00** |
| | Worst | 3.00E+00 | 8.40E+01 | 3.00E+00 | 3.00E+00 | 3.00E+00 | 3.00E+00 | 3.00E+00 | 8.43E+01 | 3.00E+00 | 3.00E+00 | 3.00E+00 | 3.00E+00 | 3.21E+01 | **3.00E+00** |
| | Mean | 3.00E+00 | 1.06E+01 | 3.00E+00 | 3.00E+00 | 3.00E+00 | 3.00E+00 | 3.00E+00 | 7.510+00 | 3.000+00 | 3.00E+00 | 3.00E+00 | 3.00E+00 | 8.90E+00 | **3.00E+00** |
| | STD | 0.00E+00 | 1.72E+01 | 3.93E-05 | 0.00E+00 | 2.29E-04 | 0.00E+00 | 0.00E+00 | 1.60E+01 | 0.000+00 | 2.74E-12 | 4.20E-15 | 2.78E-15 | 7.24E+00 | **1.74E-04** |
| F19 | Best | -3.86E+00 | -3.86E+00 | -3.86E+00 | -3.86E+00 | -3.86E+00 | -3.86E+00 | -3.86E+00 | -3.86E+00 | -3.86E+00 | -3.86E+00 | -3.86E+00 | -3.86E+00 | -3.84E+00 | **-3.00E-01** |
| | Worst | -3.86E+00 | -3.32E+00 | -3.85E+00 | -3.86E+00 | -3.79E+00 | -3.86E+00 | -3.86E+00 | -3.86E+00 | -3.86E+00 | -3.86E+00 | -3.86E+00 | -3.86E+00 | -3.13E+00 | **-3.00E-01** |
| | Mean | -3.86E+00 | -3.80E+00 | -3.86E+00 | -3.86E+00 | -3.85E+00 | -3.86E+00 | -3.86E+00 | -3.86E+00 | -3.86E+00 | -3.86E+00 | -3.86E+00 | -3.86E+00 | -3.59E+00 | **-3.00E-01** |
| | STD | 9.16E-10 | 1.05E-01 | 2.45E-03 | 3.48E-08 | 1.60E-02 | 2.63E-15 | 1.44E-03 | 3.42E-15 | 2.71E-15 | 5.26E-10 | 2.31E-15 | 2.25E-15 | 2.05E-01 | **0.00E+00** |
| F20 | Best | -3.32E+00 | -3.29E+00 | -3.32E+00 | -3.32E+00 | -3.32E+00 | -3.32E+00 | -3.32E+00 | -3.32E+00 | -3.32E+00 | -3.32E+00 | -3.32E+00 | -3.32E+00 | -2.55E+00 | **-2.73E+00** |
| | Worst | -3.20E+00 | -1.30E+00 | -3.09E+00 | -3.20E+00 | -3.07E+00 | -3.20E+00 | -3.14E+00 | -3.20E+00 | -3.20E+00 | -3.15E+00 | -3.32E+00 | -3.20E+00 | -1.08E+00 | **-1.67E+00** |
| | Mean | -3.31E+00 | -2.77E+00 | -3.27E+00 | -3.30E+00 | -3.19E+00 | -3.29E+00 | -3.22E+00 | -3.28E+00 | -3.29E+00 | -3.23E+00 | -3.32E+00 | -3.31E+00 | -1.82E+00 | **-2.34E+00** |
| | STD | 3.02E-02 | 4.31E-01 | 7.55E-02 | 4.43E-02 | 9.04E-02 | 5.34E-02 | 5.20E-02 | 5.83E-02 | 5.11E-02 | 6.33E-02 | 1.61E-15 | 3.03E-02 | 3.92E-01 | **4.44E-01** |

| No. | | AVOA | PSO | GWO | FFA | WOA | TLBO | MFO | BBO | DE | SSA | GSA | IPO | SHMS | **FWSC** |
|---|---|---|---|---|---|---|---|---|---|---|---|---|---|---|---|
| F21 | Best | -1.02E+01 | -1.01E+01 | -1.02E+01 | -1.02E+01 | -1.02E+01 | -1.02E+01 | -1.02E+01 | -1.02E+01 | -1.02E+01 | -1.02E+01 | -1.02E+01 | -1.02E+01 | -1.69E+00 | **-5.05E+00** |
| | Worst | -1.02E+05 | -6.87E-01 | -2.68E+00 | -4.81E+00 | -2.63E+00 | -2.68E+00 | -2.63E+00 | -2.63E+00 | -2.68E+00 | -2.63E+00 | -2.63E+00 | -2.63E+00 | -9.28E-01 | **-2.91E+00** |
| | Mean | -1.02E+01 | -3.78E+00 | -8.15E+00 | -9.02E+00 | -7.66E+00 | -9.26E+00 | -5.56E+00 | -5.07E+00 | -9.40E+00 | -6.30E+00 | -7.03E+00 | -8.07E+00 | -1.25E+00 | **-3.83E+00** |
| | STD | 1.08E-10 | 2.86E+00 | 2.95E+00 | 1.96E+00 | 2.90E+00 | 2.02E+00 | 3.41E+00 | 3.24E+00 | 1.99E+00 | 3.53E+00 | 3.68E+00 | 3.29E+00 | 2.42E-01 | **7.70E-01** |
| F22 | Best | -1.04E+01 | -1.04E+01 | -1.04E+01 | -1.04E+01 | -1.04E+01 | -1.04E+01 | -1.04E+01 | -1.04E+01 | -1.04E+01 | -1.04E+01 | -1.04E+01 | -1.04E+01 | -3.49E+00 | **-5.07E+00** |
| | Worst | -1.04E+01 | -1.60E+00 | -1.04E+01 | -3.68E+00 | -2.76E+00 | -3.61E+00 | -2.75E+00 | -2.75E+00 | -2.75E+00 | -2.75E+00 | -2.77E+00 | -2.75E+00 | -9.40E-01 | **-4.17E+00** |
| | Mean | -1.04E+01 | -5.04E+00 | -1.04E+01 | -9.68E+00 | -7.79E+00 | -8.70E+00 | -9.05E+00 | -5.96E+00 | -9.85E+00 | -8.89E+00 | -9.79E+00 | -9.89E+00 | -1.34E+00 | **-4.59E+00** |
| | STD | 9.14E-11 | 2.92E+00 | 1.32E-03 | 2.05E+00 | 3.09E+00 | 2.67E+00 | 2.78E+00 | 3.47E+00 | 1.85E+00 | 2.850E+00 | 1.90E+00 | 1.94E+00 | 5.06E-01 | **2.18E-01** |
| F23 | Best | -1.05E+01 | -1.05E+01 | -1.05E+01 | -1.05E+01 | -1.05E+01 | -1.05E+01 | -1.05E+01 | -1.05E+01 | -1.05E+01 | -1.05E+01 | -1.05E+01 | -1.05E+01 | -1.94E+00 | **-5.11E+00** |
| | Worst | -1.05E+01 | -1.63E+00 | -2.42E+00 | -2.87E+00 | -7.89E-01 | -3.84E+00 | -2.43E+00 | -1.86E+00 | -2.87E+00 | -2.43E+00 | -2.43E+00 | -2.42E+00 | -9.80E-01 | **-2.77E+00** |
| | Mean | -1.05E+01 | -4.89E+00 | -1.03E+01 | -9.81E+00 | -6.81E+00 | -9.87E+00 | -8.68E+00 | -5.28E+00 | -1.03E+01 | -8.53E+00 | -9.49E+00 | -7.81E+00 | -1.35E+00 | **-4.01E+00** |
| | STD | 4.34E-11 | 3.20E+00 | 1.48E+00 | 2.14E+00 | 3.34E+00 | 2.04E+00 | 3.16E+00 | 3.55E+00 | 1.40E+00 | 3.42E+00 | 2.72E+00 | 3.49E+00 | 2.43E-01 | **6.27E-01** |

**Table 9 Friedmann test for benchmark functions (F1-F13), with 30 dimensions**

| Algorithm | AVOA | PSO | GWO | FFA | WOA | TLBO | MFO | BBO | DE | SSA | GSA | IPO | SHMS | **FWSC** |
|---|---|---|---|---|---|---|---|---|---|---|---|---|---|---|
| Mean Values | 2.5385 | 13.1538 | 6.3462 | 7.5385 | 6.2308 | 4.3077 | 11.4615 | 9.3846 | 9.2308 | 9.5385 | 8.0385 | 8.3846 | 4.1923 | 4.6538 |
| Ranking | **1** | **14** | **6** | **7** | **5** | **3** | **13** | **11** | **10** | **12** | **8** | **9** | **2** | **4** |

**Table 10 Friedmann test for benchmark functions (F1-F13), with 100 dimensions**

| Algorithm | AVOA | PSO | GWO | FFA | WOA | TLBO | MFO | BBO | DE | SSA | GSA | IPO | SHMS | **FWSC** |
|---|---|---|---|---|---|---|---|---|---|---|---|---|---|---|
| Mean Values | 2.2692 | 11.6154 | 5.4615 | 11.2308 | 4.6538 | 4.2308 | 12.6923 | 7.5385 | 11.7692 | 8.9231 | 9.6923 | 7.3846 | 3.1923 | 4.3462 |
| Ranking | **1** | **11** | **6** | **12** | **5** | **3** | **14** | **8** | **13** | **8** | **10** | **7** | **2** | **4** |

**Table 11 Friedmann test for benchmark functions (F1-F13), with 500 dimensions**

| Algorithm | AVOA | PSO | GWO | FFA | WOA | TLBO | MFO | BBO | DE | SSA | GSA | IPO | SHMS | **FWSC** |
|---|---|---|---|---|---|---|---|---|---|---|---|---|---|---|
| Mean Values | 3.0385 | 8.9231 | 5.8462 | 12.0769 | 3.9615 | 4.1538 | 12.7692 | 9.0769 | 12.1538 | 8.9231 | 8.8462 | 7.8462 | 3 | 4.3846 |
| Ranking | **2** | **9** | **6** | **11** | **3** | **4** | **13** | **10** | **12** | **9** | **8** | **7** | **1** | **5** |

**Table 12 Friedmann test for benchmark functions (F1-F13), with 1000 dimensions**

| Algorithm | AVOA | PSO | GWO | FFA | WOA | TLBO | MFO | BBO | DE | SSA | GSA | IPO | SHMS | **FWSC** |
|---|---|---|---|---|---|---|---|---|---|---|---|---|---|---|
| Mean Values | 2.1538 | 9.0385 | 5.9231 | 11.7692 | 4 | 4.6538 | 12.3462 | 10.5769 | 12.7692 | 8.6923 | 8.5385 | 7.3846 | 3 | 4.1538 |
| Ranking | **1** | **10** | **6** | **12** | **3** | **5** | **13** | **11** | **14** | **9** | **8** | **7** | **2** | **4** |

**Table 13 Friedmann test for benchmark functions (F14-F23)**

| Algorithm | AVOA | PSO | GWO | FFA | WOA | TLBO | MFO | BBO | DE | SSA | GSA | IPO | SHMS | **FWSC** |
|---|---|---|---|---|---|---|---|---|---|---|---|---|---|---|
| Mean Values | 6.5 | 8.25 | 6.15 | 6.75 | 6.5 | 6.15 | 7.2 | 6.9 | 6.5 | 6.5 | 7.35 | 6.15 | 13.05 | 10.95 |
| Ranking | **2** | **7** | **1** | **3** | **2** | **1** | **5** | **4** | **2** | **2** | **6** | **1** | **9** | **8** |

Table 14 Pairwise Wilcoxon signed rank test for benchmark functions (F1-13), with 30 dimensions

| Other Algorithms vs FWSC | p-value | T+ | T- | Winner |
|---|---|---|---|---|
| AVOA vs FWSC | 0.2402 | 47 | 19 | **ANOVA** |
| PSO vs FWSC | 0.0171 | 12 | 79 | **FWSC** |
| GWO vs FWSC | 0.6772 | 33 | 45 | **FWSC** |
| FFA vs FWSC | 0.3054 | 30 | 61 | **FWSC** |
| WOA vs FWSC | 0.5771 | 26 | 40 | **FWSC** |
| TLBO vs FWSC | 0.8984 | 35 | 31 | **TLBO** |
| MFO vs FWSC | 0.0215 | 13 | 78 | **FWSC** |
| BBO vs FWSC | 0.0574 | 18 | 73 | **FWSC** |
| DE vs FWSC | 0.2734 | 29 | 62 | **FWSC** |
| SSA vs FWSC | 0.0681 | 19 | 72 | **FWSC** |
| GSA vs FWSC | 0.2439 | 63 | 28 | **FWSC** |
| IPO vs FWSC | 0.1294 | 19 | 59 | **FWSC** |
| SNAIL vs FWSC | 0.9453 | 19 | 17 | **SNAIL** |

Table 15 Pairwise Wilcoxon signed rank test for benchmark functions (F1-13), with 100 dimensions

| Other Algorithms vs FWSC | p-value | T+ | T- | Winner |
|---|---|---|---|---|
| AVOA vs FWSC | 0.1748 | 49 | 17 | **ANOVA** |
| PSO vs FWSC | 0.0076 | 9 | 82 | **FWSC** |
| GWO vs FWSC | 0.5759 | 37 | 54 | **FWSC** |
| FFA vs FWSC | 0.0044 | 7 | 84 | **FWSC** |
| WOA vs FWSC | 0.8789 | 35 | 31 | **WOA** |
| TLBO vs FWSC | 0.8125 | 36 | 30 | **TLBO** |
| MFO vs FWSC | 0.0044 | 7 | 84 | **FWSC** |
| BBO vs FWSC | 0.0078 | 9 | 82 | **FWSC** |
| DE vs FWSC | 0.0059 | 8 | 83 | **FWSC** |
| SSA vs FWSC | 0.0098 | 10 | 81 | **FWSC** |
| GSA vs FWSC | 0.1 | 10 | 81 | **FWSC** |
| IPO vs FWSC | 0.0164 | 12 | 79 | **FWSC** |
| SNAIL vs FWSC | 0.2031 | 34 | 11 | **SNAIL** |

Table 16 Pairwise Wilcoxon signed rank test for benchmark functions (F1-13), with 500 dimensions

| Other Algorithms vs FWSC | p-value | T+ | T- | Winner |
|---|---|---|---|---|
| AVOA vs FWSC | 0.168 | 49 | 17 | **ANOVA** |
| PSO vs FWSC | 0.0076 | 9 | 82 | **FWSC** |
| GWO vs FWSC | 0.5759 | 37 | 54 | **FWSC** |
| FFA vs FWSC | 0.8789 | 35 | 31 | **FFA** |
| WOA vs FWSC | 0.8789 | 35 | 31 | **WOA** |
| TLBO vs FWSC | 0.8125 | 36 | 30 | **TLBO** |
| MFO vs FWSC | 0.0044 | 7 | 84 | **FWSC** |
| BBO vs FWSC | 0.0078 | 9 | 82 | **FWSC** |
| DE vs FWSC | 0.0059 | 8 | 83 | **FWSC** |

| Other Algorithms vs FWSC | p-value | T+ | T- | Winner |
|---|---|---|---|---|
| SSA vs FWSC | 0.0098 | 10 | 81 | **FWSC** |
| GSA vs FWSC | 0.01 | 10 | 81 | **FWSC** |
| IPO vs FWSC | 0.0164 | 12 | 79 | **FWSC** |
| SNAIL vs FWSC | 0.2031 | 34 | 11 | **SNAIL** |

**Table 17 Pairwise Wilcoxon signed rank test for benchmark functions (F1-13), with 1000 dimensions**

| Other Algorithms vs FWSC | p-value | T+ | T- | Winner |
|---|---|---|---|---|
| AVOA vs FWSC | 0.123 | 51 | 15 | **ANOVA** |
| PSO vs FWSC | 0.0061 | 8 | 83 | **FWSC** |
| GWO vs FWSC | 0.0479 | 17 | 74 | **FWSC** |
| FFA vs FWSC | 0.0034 | 6 | 85 | **FWSC** |
| WOA vs FWSC | 0.9658 | 34 | 32 | **WOA** |
| TLBO vs FWSC | 0.7002 | 28 | 38 | **FWSC** |
| MFO vs FWSC | 0.0034 | 6 | 85 | **FWSC** |
| BBO vs FWSC | 0.0034 | 6 | 85 | **FWSC** |
| DE vs FWSC | 0.0034 | 6 | 85 | **FWSC** |
| SSA vs FWSC | 0.0081 | 9 | 82 | **FWSC** |
| GSA vs FWSC | 0.0081 | 9 | 82 | **FWSC** |
| IPO vs FWSC | 0.0171 | 12 | 79 | **FWSC** |
| SNAIL vs FWSC | 0.2031 | 34 | 11 | **SNAIL** |

**Table 18 Pairwise Wilcoxon signed rank test for benchmark functions (F14-23)**

| Other Algorithms vs FWSC | p-value | T+ | T- | Winner |
|---|---|---|---|---|
| AVOA vs FWSC | 0.0156 | 28 | 0 | **ANOVA** |
| PSO vs FWSC | 0.0312 | 27 | 1 | **PSO** |
| GWO vs FWSC | 0.0156 | 28 | 0 | **GWO** |
| FFA vs FWSC | 0.0156 | 28 | 0 | **FFA** |
| WOA vs FWSC | 0.0156 | 28 | 0 | **WOA** |
| TLBO vs FWSC | 0.0156 | 28 | 0 | **TLBO** |
| MFO vs FWSC | 0.0312 | 27 | 1 | **MFO** |
| BBO vs FWSC | 0.0312 | 27 | 1 | **BBO** |
| DE vs FWSC | 0.0156 | 28 | 0 | **DE** |
| SSA vs FWSC | 0.0156 | 28 | 0 | **SSA** |
| GSA vs FWSC | 0.0312 | 27 | 1 | **GSA** |
| IPO vs FWSC | 0.0156 | 28 | 0 | **IPO** |
| SNAIL vs FWSC | 0.0977 | 8 | 37 | **FWSC** |

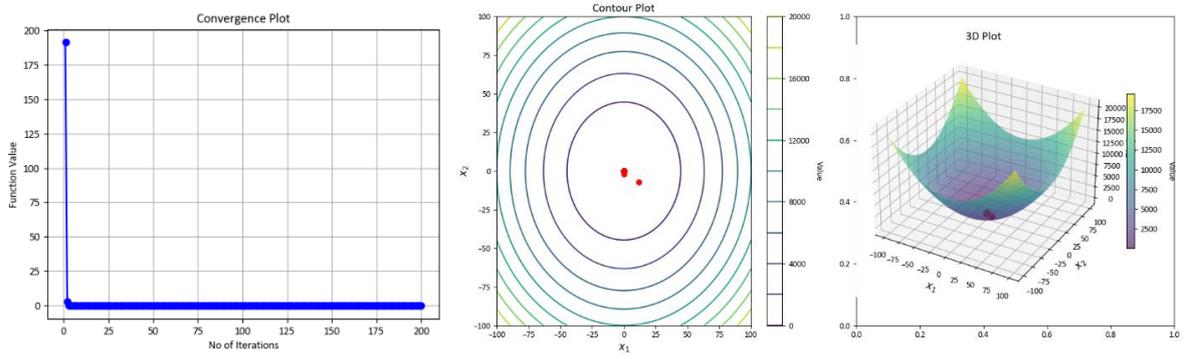

**Figure 4(a): Sphere Function**

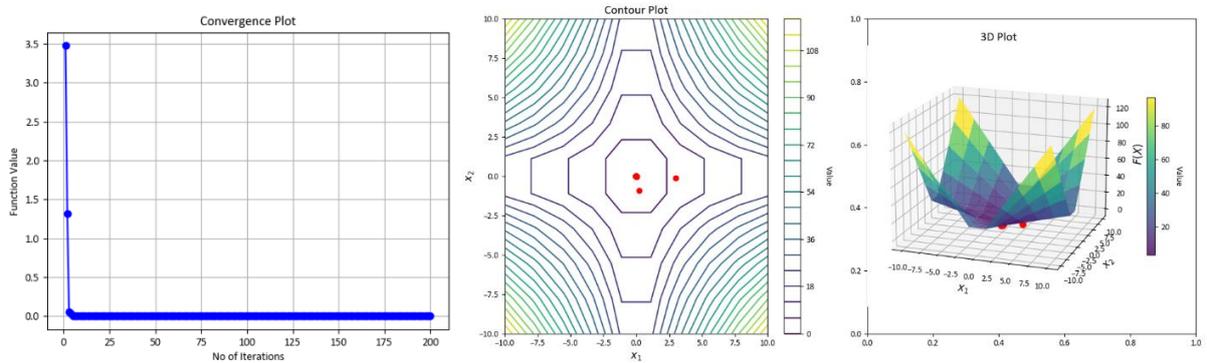

**Figure 4(b): Schwefel 2.22 Function**

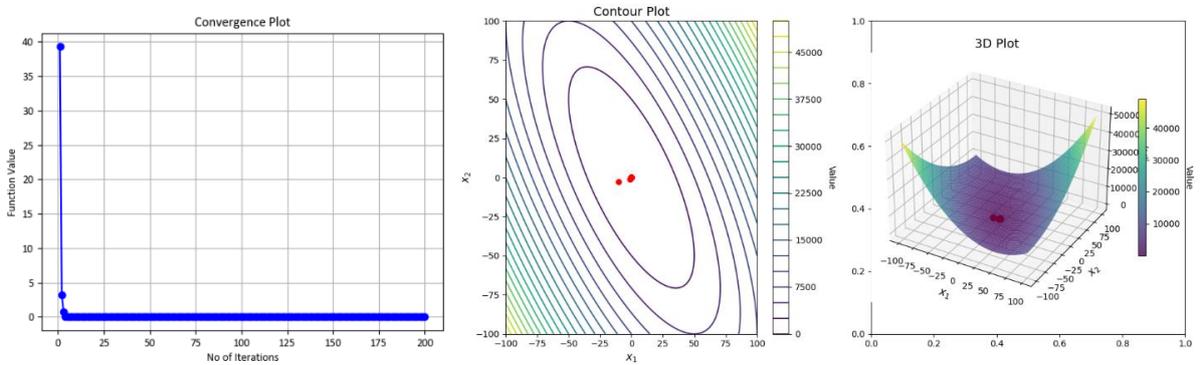

**Figure 4(c): Schwefel 1.2 Function**

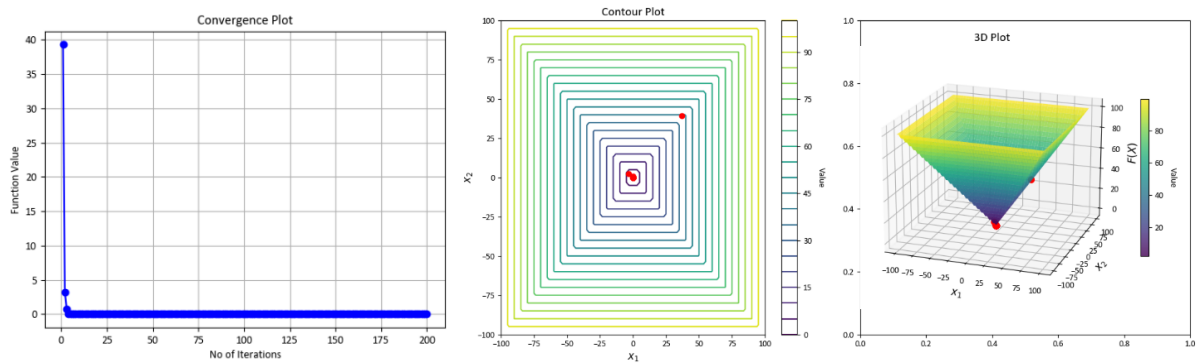

**Figure 4(d): Schwefel 2.21 Function**

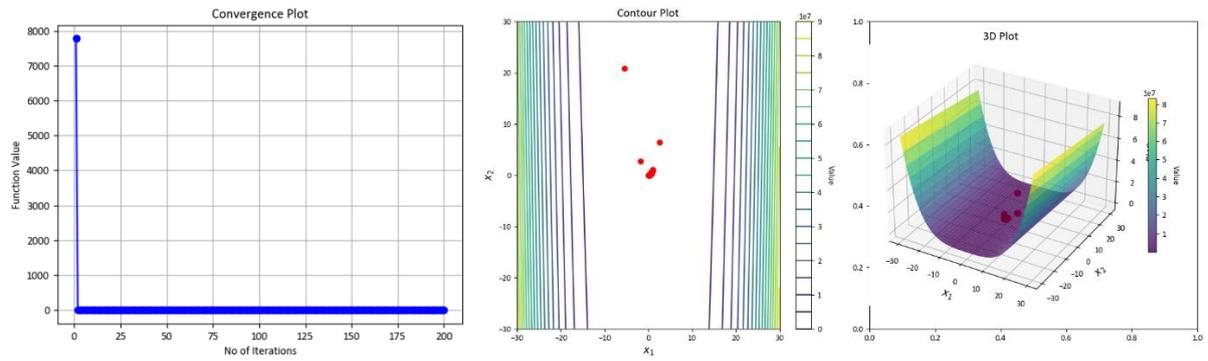

Figure 4(e): Rosenbrock Function

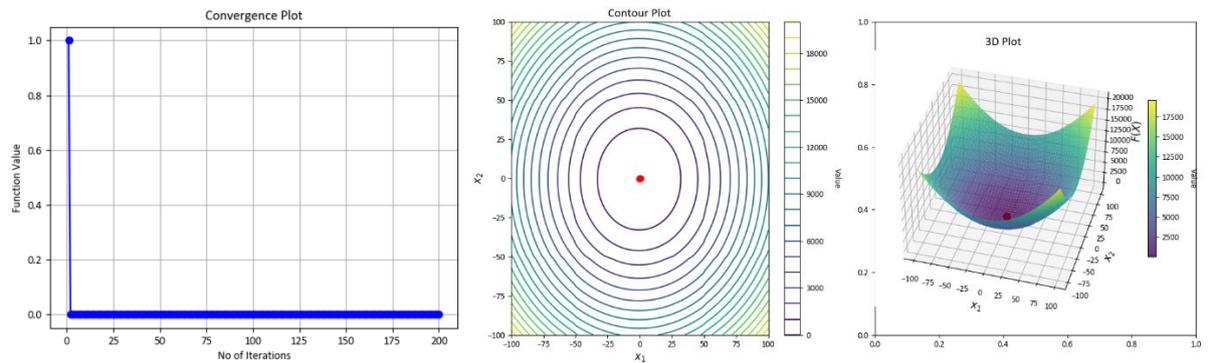

Figure 4(f): Step Function

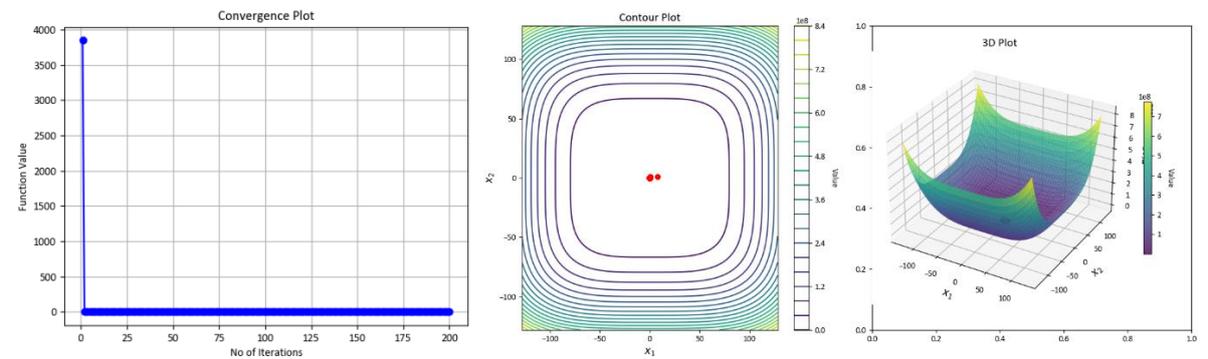

Figure 4(g): Quartic Function

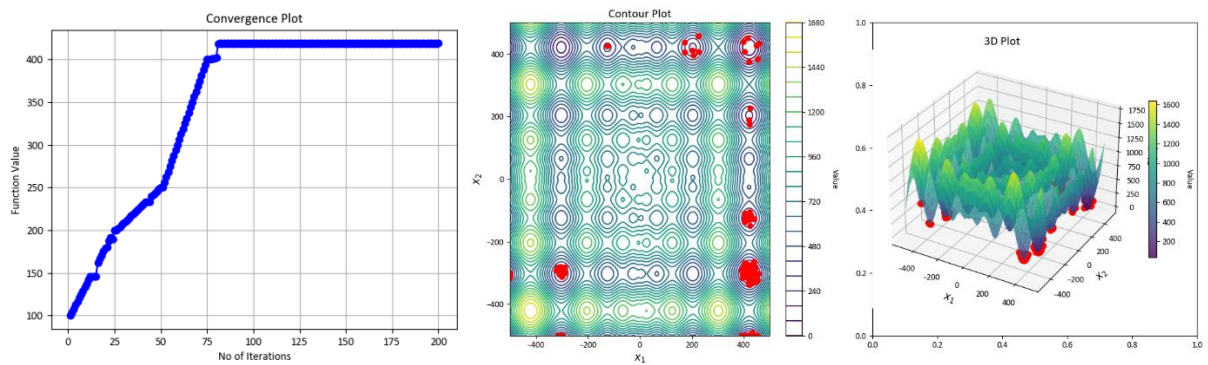

Figure 4(h): Schwefel Function

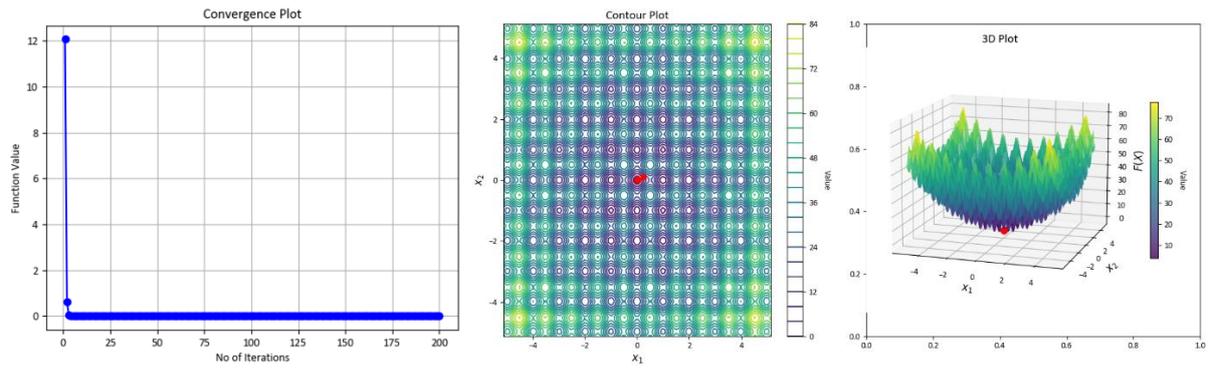

Figure 4(i): Rastrigin Function

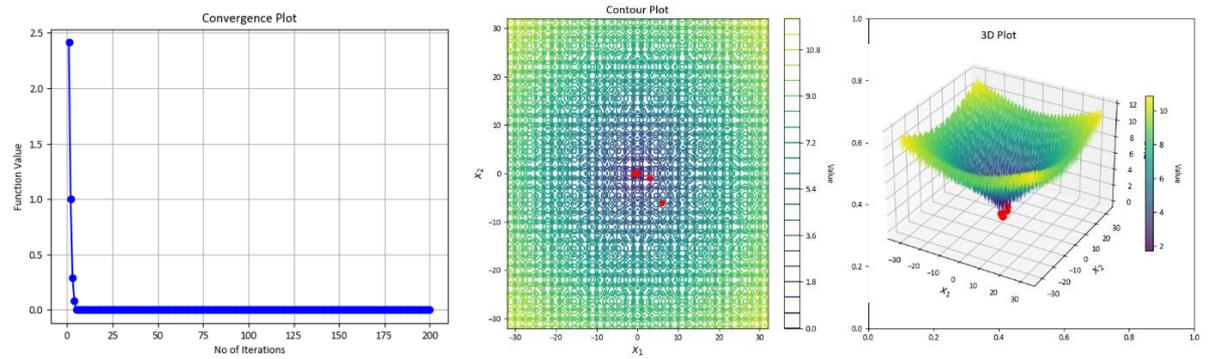

Figure 4(j): Ackley Function

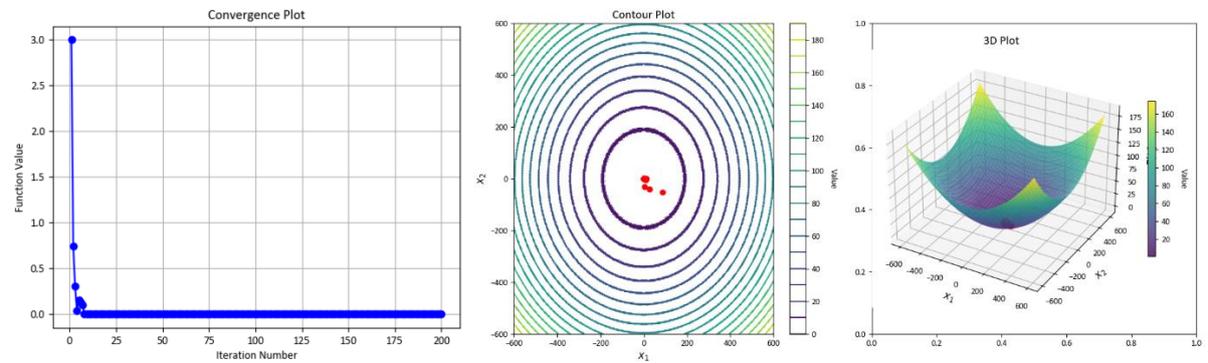

Figure 4(k): Griewank Function

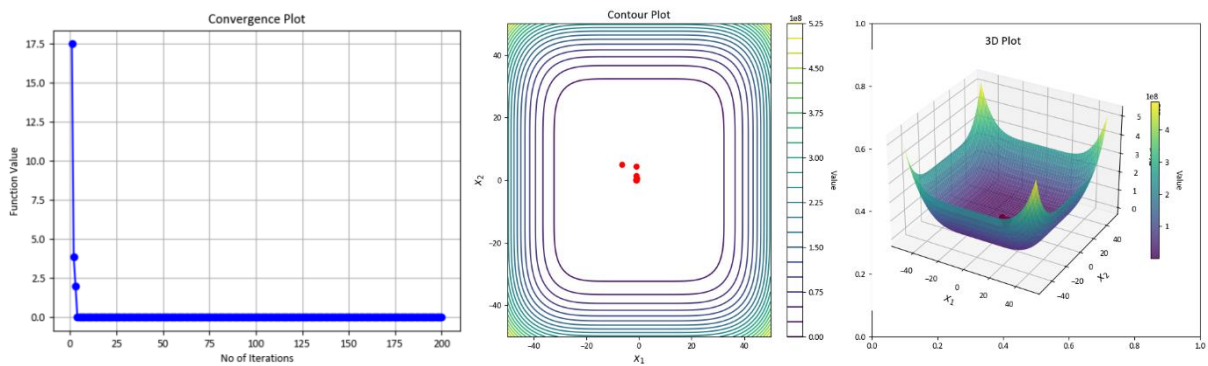

Figure 4(l): Penalized Function

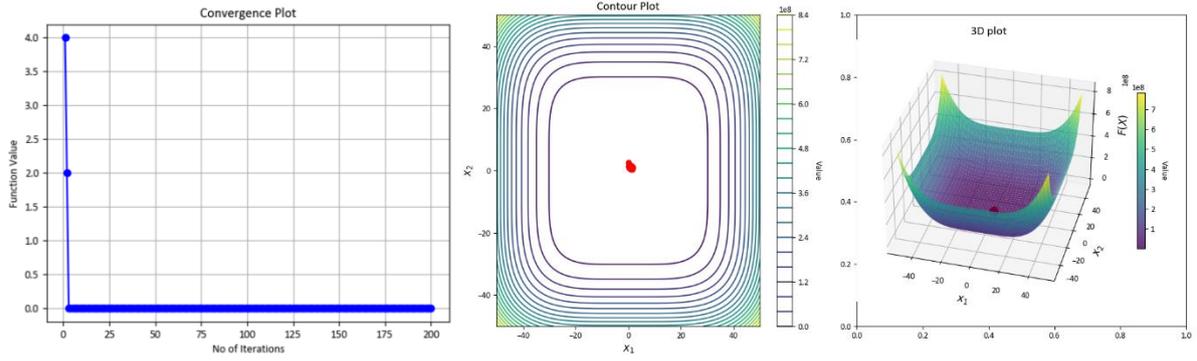

**Figure 4(m): Penalized 2 Function**

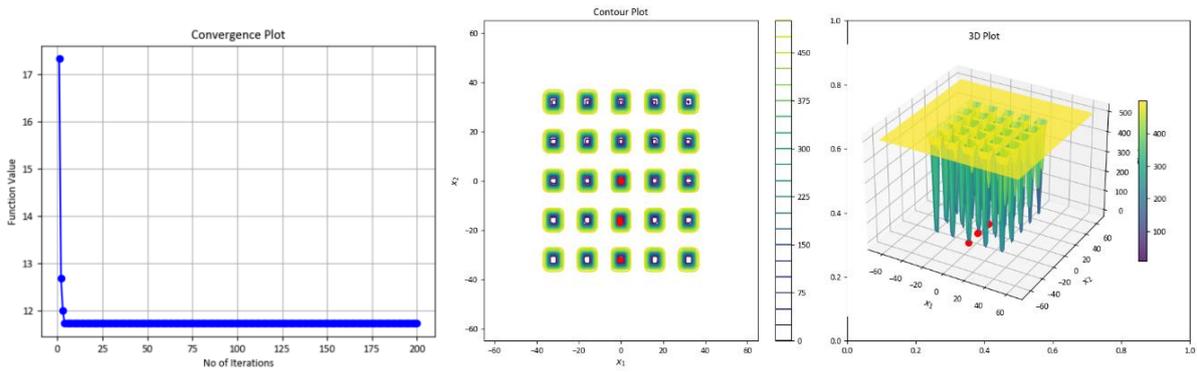

**Figure 5(a): Foxhole Function**

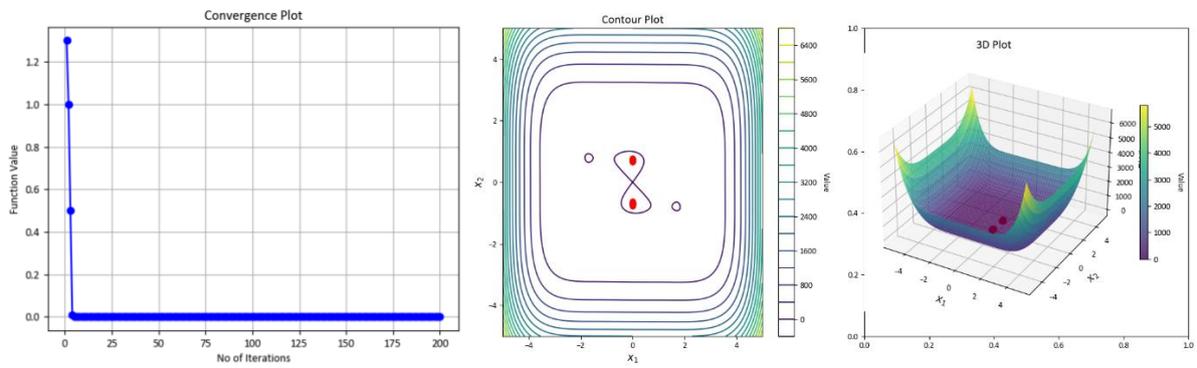

**Figure 5(b): Six Hump Camel Function**

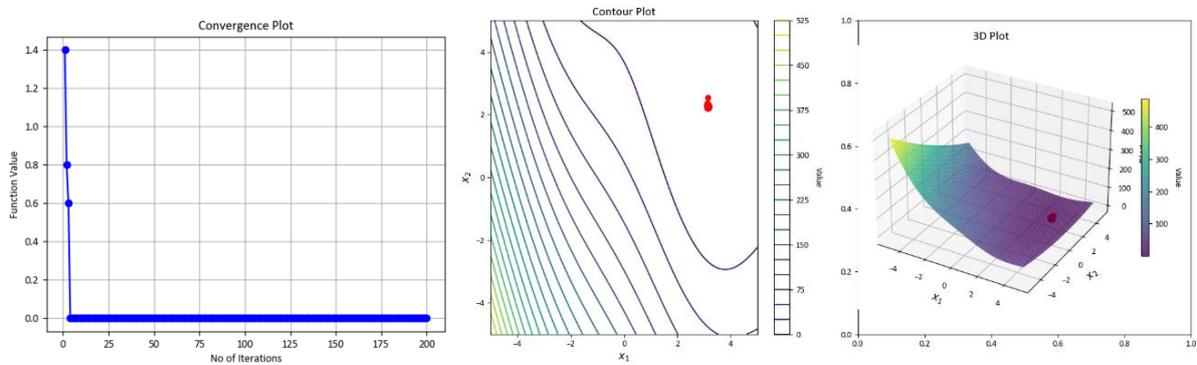

**Figure 5(c): Branin Function**

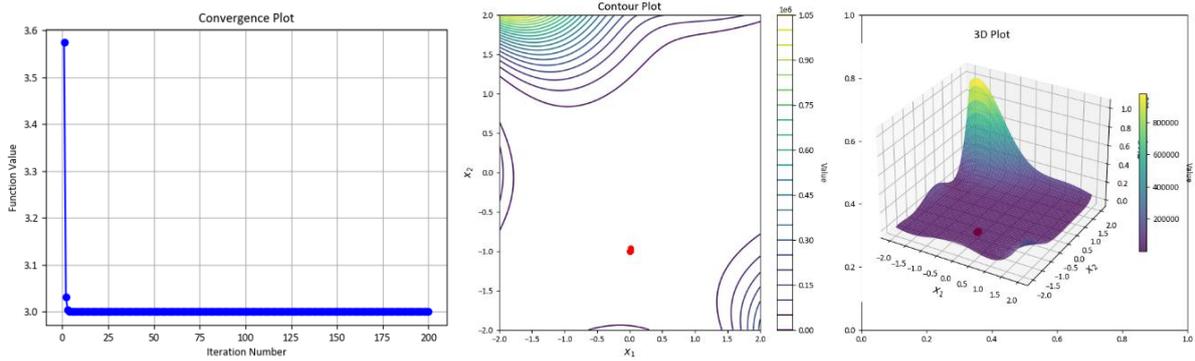

Figure 5(d): Goldstein-Price Function

## 4. Solution to real world problems

### 4.1 Pressure Vessel Design optimization problem

The pressure vessel design problem, shown in Figure 6 is originally presented by Sandgren (1990), is a classic engineering optimization task aimed at minimizing the total cost of materials, fabrication, and welding, while meeting four specific constraints. In this problem, the thickness of the shell ($T_s$) and spherical head thickness ($T_h$) are treated as discrete variables, while the inner radius of shell ($R$), and shell length ($L$) are continuous variables.

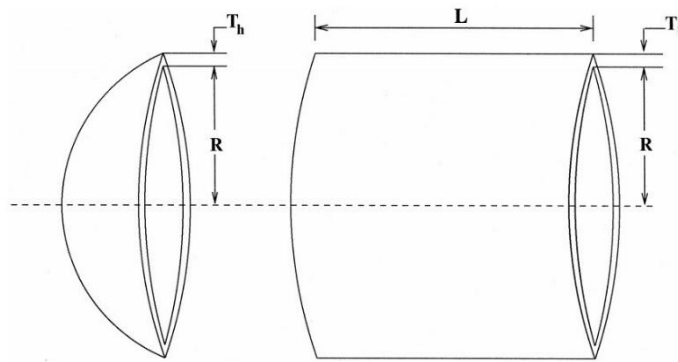

Figure 6: Tube and Pressure Vessel (Kale and Kulkarni, 2021)

This optimization problem has been tackled using various metaheuristic approaches, including Nonlinear Integer and Discrete Programming (NIDPM) by Sandgren (1990), Augmented Lagrange by Kannan and Kramer (1994), GA by Coello (2000), Chaos Particle Swarm Optimization (CPSO) by He and Wang (2007), LCA by Kashan (2011), Optics Inspired Optimization (OIO) by Kashan (2015), and Multi Random Start Local Search (MRSLS) by Kulkarni et al. (2016). The results of these comparisons are summarized in Table 9, while the convergence curves for the FWSC algorithm are displayed in Figure 7.

Table 9 Comparison of FWSC results Design and Optimization of Pressure Vessel Design Problem

| Variables | NIDPM | Augmented Lagrange | GA (2000) | CPSO | LCA | OIO | MRSLS | FWSC |
|---|---|---|---|---|---|---|---|---|
| $T_s$ | 1.125 | 1.125 | 0.8125 | 0.8125 | NA | NA | 0.8125 | **0.8750** |
| $T_h$ | 0.625 | 0.625 | 0.4375 | 0.4375 | NA | NA | 0.4375 | **0.4375** |
| $R$ | 48.97 | 58.291 | 40.3239 | 42.0912 | NA | NA | 41.9645 | **44.5025** |
| $L$ | 106.72 | 43.69 | 200 | 176.7465 | NA | NA | 178.3043 | **149.0235** |
| Cost $f(x)$ | 7981.5690 | 7198.0428 | 6288.7445 | 6061.0777 | 6059.8553 | 6059.7143 | 6076.1215 | **6189.6362** |

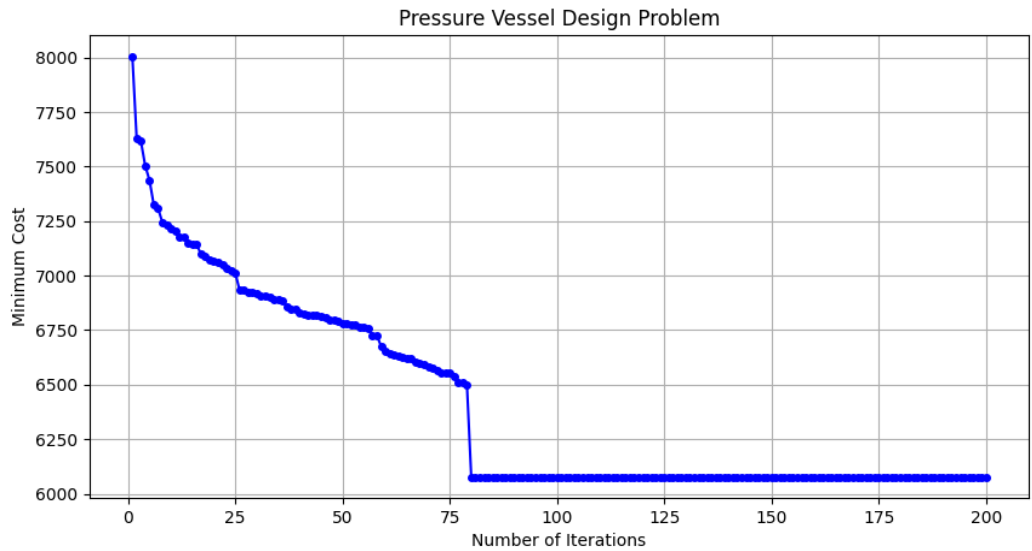

Figure 7: Convergence curve of Pressure Vessel Design Problem

## 4.2 Stepped Cantilever Beam Design optimization problem

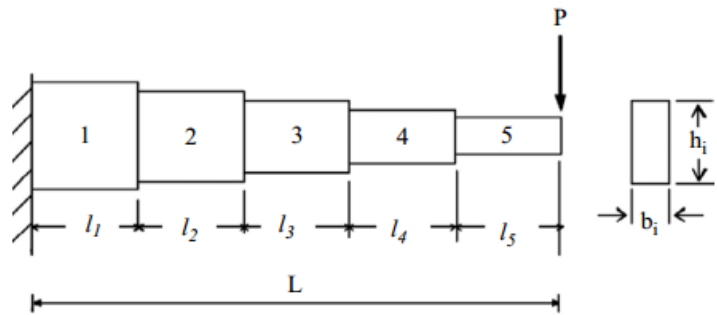

Figure 8: Stepped Cantilever Beam (Kale and Kulkarni, 2021)

The mixed-variable stepped cantilever beam design, illustrated in Figure 8 is an engineering optimization problem proposed by Thanedar and Vanderplaats (1995). The goal is to minimize the volume $V$ of a statically loaded cantilever beam made of alloyed steel. The design variables include both discrete variables, such as $(b_1, h_1, b_2, h_2, b_3, h_3)$, where $b$ represents the width and $h$ represents the height of the beam cross-section, and continuous variables like $(b_4, h_4, b_5, h_5)$.

This optimization problem has been previously solved using a variety of metaheuristic algorithms, including Branch & Bound (B&B - RU) by Thanedar and Vanderplaats (1995), Adaptive Penalty Method (GA - APM) by Lemonge and Barbosa (2004), AIS - GA - C by Bernardino (2007), Firefly Algorithm (FA) by Gandomi et al. (2011), Colliding Bodies Optimization (CBO) by Kaveh et al. (2014), MRSLS by Kulkarni et al. (2016), and Cohort Intelligence using Static Penalty Function (CI-SPF) by Kale and Kulkarni (2022). The comparative results from these methods are summarized in Table 10, with the convergence plots generated by the FWSC algorithm shown in Figure 9.

Table 10 Comparison of FWSC results Design and Optimization of Stepped Cantilever Beam Design Problem

| Variables | B&B-RU | GA-APM | AIS-GA-C | FA | CBO | MRSLS | CI-SPF | FWSC |
|---|---|---|---|---|---|---|---|---|
| $b_1$ | 4 | 3 | 3 | 3 | 4 | 3 | 3 | **3** |
| $h_1$ | 62 | 60 | 60 | 60 | 55 | 60 | 60 | **60** |
| $b_2$ | 3.1 | 3.1 | 3.1 | 3.1 | 3 | 3.1 | 3.1 | **3.1** |
| $h_2$ | 60 | 55 | 60 | 55 | 55 | 55 | 55 | **55** |
| $b_3$ | 2.6 | 2.6 | 2.6 | 2.6 | 3 | 2.6 | 2.6 | **2.6** |
| $h_3$ | 55 | 50 | 50 | 50 | 52 | 50 | 50 | **50** |
| $b_4$ | 2.2052 | 2.289 | 2.3110 | 2.2050 | 3.0458 | 2.21 | 2.2046 | **2.2896** |
| $h_4$ | 44.09 | 45.6260 | 43.168 | 44.091 | 42.3208 | 44.06 | 44.0915 | **44.2750** |
| $b_5$ | 1.751 | 1.7930 | 2.2250 | 1.75 | 2.7314 | 1.75 | 1.7497 | **1.7711** |
| $h_5$ | 35.03 | 34.5930 | 31.25 | 34.995 | 45.2605 | 34.99 | 34.9951 | **34.8105** |
| Volume ($V$) | 73555.00 | 64698.56 | 66533.47 | 63893.52 | 80329.3686 | 63903.41 | 63893.4544 | **64352.9122** |

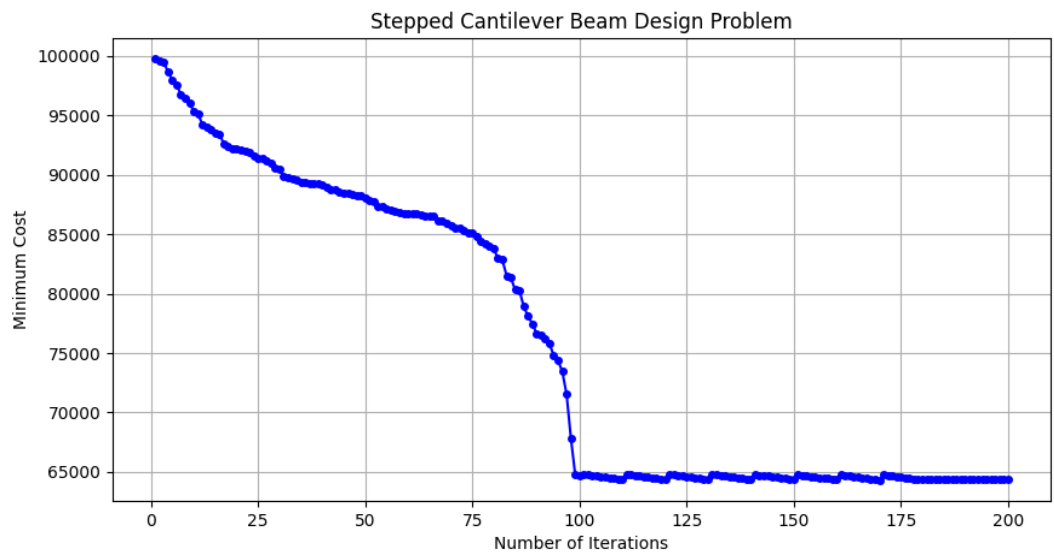

Figure 9: Convergence curve of Stepped Cantilever Beam Design Problem

## 4.3 Welded Beam Design optimization problem

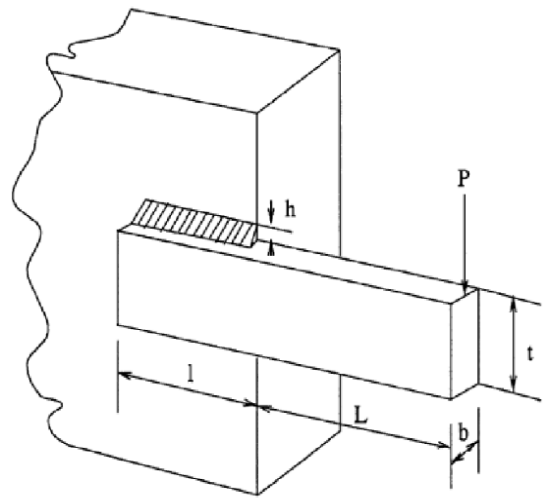

Figure 10: Welded Beam Design Problem (Kale and Kulkarni,2021)

The welded beam design problem, shown in Figure 10, was originally discussed in the works of Coello and Montes (2002), Coello (2000), and He and Wang (2007). The objective here is to minimize the cost, with four continuous design variables: $h, l, t, b$, representing the height, length, thickness, and breadth of the beam, respectively. This problem has also been solved using several metaheuristic algorithms, including GA (Deb, 1991), GA by Coello (2000), GA by Coello and Montes (2002), PSO by He and Wang (2007), CBO by Kaveh et al. (2014), CI-SPF by Kale and Kulkarni (2022), and MRSLS by Kulkarni et al. (2016). The comparison results for these methods are presented in Table 11, while the convergence plots derived from the FWSC algorithm are shown in Figure 11.

Table 11 Comparison of FWSC results Design and Optimization of Wielded Beam Design

| Variables | GA (1991) | GA (2000) | GA (2002) | PSO | CBO | CI-SPF | MRSLS | FWSC |
|---|---|---|---|---|---|---|---|---|
| $h$ | 0.2400 | 0.2088 | 0.2059 | 0.2023 | 0.2057 | 0.1974 | 0.1408 | **0.2092** |
| $l$ | 6.1730 | 3.4205 | 3.4713 | 3.5442 | 3.4704 | 3.0884 | 4.6724 | **3.4872** |
| $t$ | 8.1789 | 8.9975 | 9.0202 | 9.0482 | 9.0372 | 9.9988 | 9.8254 | **9.0936** |
| $b$ | 0.2533 | 0.21 | 0.2064 | 0.2057 | 0.2057 | 0.168 | 0.1740 | **0.2868** |
| Cost $f(x)$ | 2.4331 | 1.74831 | 1.7282 | 1.7280 | 1.7246 | 1.5587 | 1.6384 | **1.7275** |

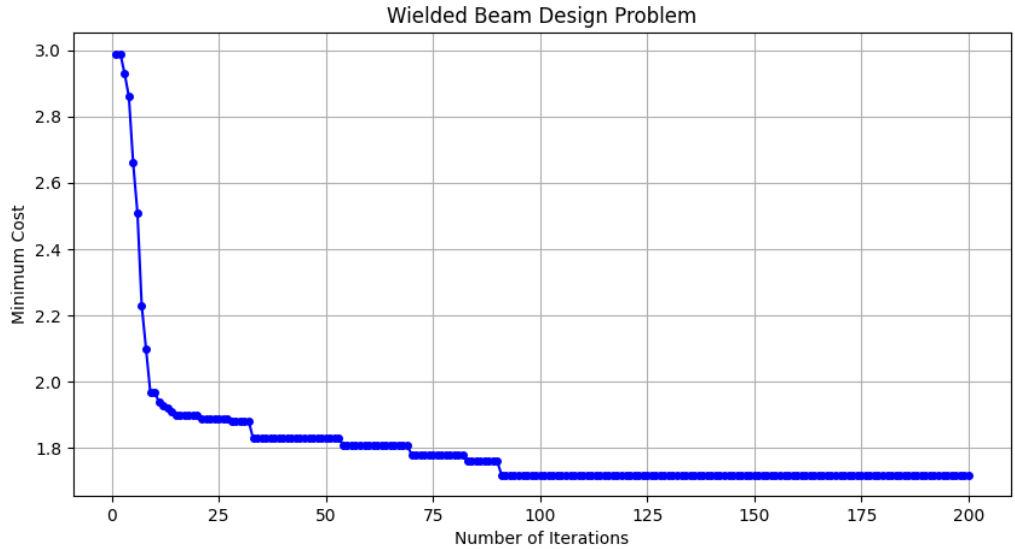

Figure 11: Convergence curve of Wielded Beam Design Problem

# 6. Conclusions and Future Directions

In this study, we introduced a novel metaheuristic optimization algorithm referred to as Fig Tree-Wasp Symbiotic Coevolutionary (FWSC) algorithm. The symbiotic coevolutionary relationships between fig trees and wasps are mathematically modeled. The mating of wasps, pollinating the figs, searching for new trees for pollination and wind effect drifting of wasps are modeled in the algorithm. The algorithm is validated by solving a set of unimodal (UM) and multi-modal (MM) test problems. Furthermore, the scalability of the FWSC algorithm is examined by solving certain problems with higher dimensions. The Statistical analyses using the Wilcoxon and Friedman tests are performed to compare the FWSC algorithm's performance against several existing optimization algorithms. The exhibited that the FWSC outperforms algorithms such as PSO, GWO, FFA, WOA, TLBO, MFO, BBO, DE, SSA, GSA, and IPO in terms of objective function across all tested dimensions. Moreover, the algorithm exhibited promising performance when applied to real-world engineering optimization problems, such as pressure vessel design, stepped cantilever beam design, and welded beam design. The solutions to these problems underscored the versatility and robustness of the FWSC algorithm in solving constrained problems.

Overall, the FWSC algorithm demonstrated a strong potential for efficiently solving complex optimization problems, both in theoretical benchmark settings and practical engineering applications. Its ability to exploit the symbiotic relationship between fig trees and wasps provides a powerful metaphor for the coevolutionary algorithm. The inherent capabilities of the algorithm for handling constraints needs to be explored which may avoid incorporation of any supporting technique. The algorithm needs to be explored for solving constrained discrete and combinatorial problems especially in the domain of resource allocation and supply chain management.

**Declarations**: Conflict of interest Authors have no competing and conflicting interests of any kind.

**Ethical approval**: This article does not contain any studies with human participants or animals performed by any of the authors.

**Contributions**:

Author 1 (Anand J Kulkarni): FWSC algorithm conceptualization, mathematical modelling, manuscript writing and review

Author 2 (Isha Purnapatre): FWSC algorithm coding, testing, manuscript writing

Author 3 (Apoorva S Shastri): Statistical analysis, manuscript writing and review